\documentclass{article}

\usepackage[utf8]{inputenc}
\usepackage{amsmath}
\usepackage{amssymb}
\usepackage{amsthm}
\usepackage{graphicx}
\usepackage{bm,bbm}

\usepackage{listings}
\usepackage[dvipsnames]{xcolor}
\usepackage[colorlinks = true,
            linkcolor = blue,
            urlcolor  = blue,
            citecolor = OliveGreen,
            anchorcolor = blue]{hyperref}
\usepackage{cleveref}
\usepackage{mleftright}
\usepackage{enumitem}
\usepackage{url}
\usepackage{algorithmic}
\usepackage[bottom]{footmisc}

\usepackage{titlesec}
\titlespacing*{\paragraph}{0pt}{1ex plus .2ex}{0.5ex}

\usepackage{caption}
\usepackage{subcaption}

\usepackage{natbib}
\usepackage{booktabs}

\usepackage{comment}

\definecolor{RoyalBlue}{RGB}{0,100,170}
\definecolor{peach}{rgb}{1, 0.56, 0.56}
\definecolor{midgray}{RGB}{150,150,150}
\definecolor{EasternBlue}{RGB}{37,150,190}
\definecolor{sand}{RGB}{250,150,120}
\definecolor{grass}{RGB}{120, 190, 50}
\definecolor{sky}{RGB}{50,150,250}
\definecolor{Orange}{RGB}{250,150,50}
\definecolor{Cerulean}{RGB}{80,150,220}
\definecolor{Emerald}{RGB}{62,156,94}
\definecolor{Rouge}{RGB}{250,95,95}
\definecolor{coral}{RGB}{240,128,128}
\definecolor{violet}{RGB}{139, 96, 240}

\definecolor{ColorDef}{RGB}{80, 180, 150}
\definecolor{RevisionRed}{RGB}{240,35,35}
\definecolor{RevisionBlue}{RGB}{80,180,250}
\definecolor{TODO}{RGB}{255,150,50}

\include{defs}
\def\shownotes{0}  
\ifnum\shownotes=1
\newcommand{\authnote}[2]{{$\ll$\textsf{\footnotesize #1 notes: #2}$\gg$}}
\else
\newcommand{\authnote}[2]{}
\fi

\usepackage{environ}

\newif\ifsolution
\solutiontrue
\newcounter{solnequation}
\newcounter{solneqtmp}
\NewEnviron{soln}{
    \setcounter{solneqtmp}{\value{equation}}
    \setcounter{equation}{\value{solnequation}}
    \renewcommand\theequation{S\arabic{equation}}
    \ifsolution\color{red}\expandafter\textbf{Solution} \BODY\fi%
    \setcounter{solnequation}{\value{equation}}
    \setcounter{equation}{\value{solneqtmp}}
    \renewcommand\theequation{\arabic{equation}}}
\usepackage{romannum}

\usepackage[main,final]{neurips_2026}
\usepackage{todonotes}
\usepackage{pgfplots}
\pgfplotsset{compat=1.16}

\makeatletter
\renewcommand{\paragraph}{%
  \@startsection{paragraph}{4}%
  {\z@}{1ex \@plus 1ex \@minus .2ex}{-1em}%
  {\normalfont\normalsize\bfseries}%
}
\makeatother

\title{Irreducible Supervision Enables Compositional Generalization in Post-Training}
\author{%
  Ellen Ma \\
  DoorDash \\
  San Francisco, CA, USA \\
  \texttt{ellen.ma@doordash.com} \\
  \And
  Nikhil Anand \\
  Kempner Institute, Harvard University \\
  Cambridge, MA, USA
}
\date{}

\newcommand{\N}{\mathbb N}

\newtheorem*{namedtheorem}{\theoremname}
\newcommand{\theoremname}{testing}

\newtheorem*{theorem*}{Theorem}

\newtheorem*{lemma*}{Lemma}

\newtheorem*{question*}{Question}

\theoremstyle{definition}

\newtheorem*{definition*}{Definition}

\newtheorem*{remark*}{Remark}

\theoremstyle{plain}

\def\eqref#1{equation~\ref{#1}}

\def\1{\bm{1}}

\DeclareMathAlphabet{\mathsfit}{\encodingdefault}{\sfdefault}{m}{sl}
\SetMathAlphabet{\mathsfit}{bold}{\encodingdefault}{\sfdefault}{bx}{n}

\newcommand{\R}{\mathbb{R}}

\begin{document}
\maketitle
\pagenumbering{arabic}

\def \R {\mathbb{R}}
\def \N {\mathcal{N}}

\begin{abstract}
    Reinforcement learning (RL) with verifiable rewards has driven recent gains in large language model (LLM) reasoning. However, whether it creates new capabilities or sharpens existing ones remains debated. We study this question for compositional generalization in a controlled program synthesis domain where compositional depth is well-defined, shortcuts are mechanically detectable, and the base model has zero capability before fine-tuning. We find that compositional capability is established during SFT, and that it depends on the supervised fine-tuning (SFT) data being \textit{irreducible} at two levels: the training programs must not collapse to shallower equivalents (structural irreducibility), and the input-output examples shown in the prompt must rule out shallower solutions (observational irreducibility). Neither alone is sufficient. Even moderate contamination with reducible exemplars degrades performance, and full contamination leaves performance at the noise floor. Given this SFT foundation, standard outcome-reward RL (GRPO, PPO) maintains this but does not extend compositional performance, contrasting with prior results in simpler nested function execution settings. Results hold on Qwen3-8B, replicate within-family at 4B and 32B, and replicate cross-family on Llama-3.1-8B. In this setting, the bottleneck for compositional capability is the structure of the SFT data, set before RL begins. Code release pending institutional IP review.
\end{abstract}

\section{Introduction}
\label{sec:intro}

Recent progress in large language model (LLM) reasoning has largely been driven by post-training~\citep{Ouyang2022TrainingLM}. The typical pipeline is supervised fine-tuning (SFT) on demonstrations followed by reinforcement learning (RL) against verifiable rewards~\citep{DeepSeekAI2025DeepSeekR1IR}. However, how post-training leads to true skill mastery is debated. One line of work argues that RL drives the acquisition of new capabilities~\citep{Yuan2025FromFA, Liu2025ProRLPR}; another that RL only amplifies correct traces already present in the pre-training or SFT distributions~\citep{Yue2025DoesRL,Foster2025IsAG,Zhao2025EchoCR}. We identify an upstream variable that shapes this debate: compositional capability is established during SFT, and whether it emerges at all depends on structural properties of the SFT data, before RL begins.

We isolate the source of compositional generalization across SFT and RL in a controlled program synthesis domain where compositional depth is well-defined and shortcuts are detectable. The task is to produce a program in \texttt{arc-dsl}, a functional language of 119 typed primitives over integer grids, given three input-output (I/O) examples. The central property we manipulate is whether a training task genuinely requires composition, or instead admits shortcut solutions. A depth-$d$ program is \textit{structurally reducible} if a shallower program computes the same function, and \textit{structurally irreducible} otherwise. Separately, a prompt is \textit{observationally reducible} if the three visible I/O pairs happen to be consistent with some shallower program, even when the underlying program is not. Figure~\ref{fig:dsl} illustrates the task.

Summarizing our contributions:

\begin{itemize}
    \item We formalize irreducibility at two levels (structural: no shallower DSL program is equivalent, and observational: no shallower-program solution to the prompt I/O pairs exists) and provide a filtering pipeline that enforces both.
    \item We show that compositional generalization from SFT requires irreducibility at both the structural and observational levels and neither alone suffices.
    \item We show that standard outcome-reward RL (GRPO, PPO) from a non-CoT SFT initialization maintains but does not extend compositional performance, contrasting with results in nested-function-execution settings~\citep{Yuan2025FromFA}. A log-probability analysis shows RL narrows the correct-incorrect discrimination gap under sparse reward rather than sharpening onto correct solutions.
    \item We characterize the limits of compositional transfer: depth-$d$ generalization requires direct depth-$d$ supervision, with partial transfer to shallower depths and weak transfer to deeper ones.
\end{itemize}
 
\section{Methods}

\subsection{Task domain: DSL program synthesis over grid transformations}
\label{sec:task-domain}

\begin{figure*}
    \centering
    \includegraphics[width=\textwidth]{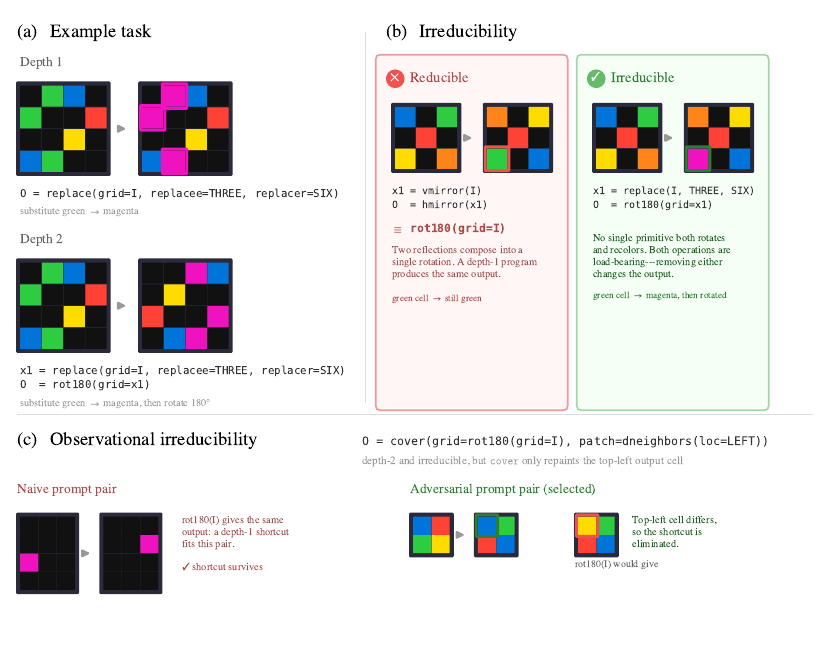}
    \caption{Task domain and irreducibility. (a) Example tasks at depth 1 and depth 2; the model receives I/O pairs and must produce the program. (b) A depth-2 program is structurally reducible if a single primitive computes the same function (left: \texttt{vmirror}$\circ$\texttt{hmirror} = \texttt{rot180}) and structurally irreducible otherwise (right: no single primitive both recolors and rotates). (c) Observational irreducibility: for the depth-2 program shown, a naive prompt pair is also explained by the depth-1 shortcut \texttt{rot180} (the \texttt{cover} step repaints only the top-left output cell, already background there), while an adversarially selected pair (a $2{\times}2$ grid) separates the two programs. Grids and program were produced by our task generator. See Appendix~\ref{app:dsl} for details.}
    \label{fig:dsl}
\end{figure*}

We study compositional skill emergence in program synthesis over \texttt{arc-dsl}~\citep{Hodel2024rearc}\footnote{\url{https://github.com/michaelhodel/arc-dsl}}, a domain-specific language of 119 typed primitives operating on integer grids (color modification, spatial operations, etc). A \texttt{Grid} is the two-dimensional integer array used as task input and output; non-\texttt{Grid} types are intermediate values such as \texttt{Integer}, \texttt{IntegerTuple}, and \texttt{Objects} (connected components). For example, \texttt{rot90(I)} returns a \texttt{Grid} and \texttt{width(I)} returns an \texttt{Integer}, so \texttt{hupscale(rot90(I), width(I))} is type-valid because \texttt{hupscale} accepts a \texttt{Grid} and an \texttt{Integer}. Programs are sampled at controlled compositional depth $d \in\{ 1,\ldots,5\}$, with a branching constraint that prevents degenerate linear chains. Each task presents 3 input-output grid pairs and correctness is evaluated on 22 additional held-out pairs (Appendix~\ref{app:dsl}). Evaluation uses 500 held-out tasks per depth. See Figure~\ref{fig:dsl} for examples. 

\paragraph{Structural irreducibility guarantees.}

Formally, a depth-$d$ program $q=f_d\circ\dots\circ f_1$ is structurally irreducible if there does not exist a program $r=g_p\circ\dots\circ g_1$ such that $p<d$ and $r(x) = q(x)$ on a validation pool $\mathcal{G}$, a fixed set of 29 diagnostic grids (defined below) on which behavioral equivalence is tested. Without this irreducibility, apparent depth-$d$ performance may reflect shallow pattern matching (Section~\ref{sec:res-irreducible}).

We enforce structural irreducibility through a multi-layer pipeline combining construction-time inverse-pair filters with verification-time behavioral fingerprinting against an exhaustive enumeration of depth-1 programs and a sampled cache of higher-depth programs. We validate the pipeline empirically (Appendix~\ref{app:irreducibility}), bounding residual contamination below $\approx$1.5\%. Evaluation data is strengthened so depth-1 shortcuts are structurally impossible; deeper shortcuts are bounded by cache coverage. For full details see Appendix~\ref{app:dsl}. 

\paragraph{Observational irreducibility via adversarial I/O pair selection.} Structural irreducibility is necessary but not sufficient. With only 3 input-output pairs in the prompt, many depth-2 tasks are observationally reducible. To distinguish a program from shallower solutions, we select prompt I/O pairs adversarially in two phases. First, we run the target program on 29 diagnostic grids chosen to break symmetry assumptions (dimension extremes, uniform color, checkerboard, diagonal). Second, we greedily select the final 3 prompt pairs by set-cover over this pool to eliminate as many depth-1 shortcuts as possible across the three prompt pairs (Appendix~\ref{app:irreducibility}). For example, if every prompt grid happens to be symmetric under a mirror primitive, then composing that mirror with a recoloring produces the same outputs as the recoloring alone, so the prompt cannot distinguish the depth-2 program from a depth-1 shortcut; the adversarial selection instead picks a non-symmetric grid on which the two programs differ.

\subsection{Training setup}
\label{sec:training-setup}

For full hyperparameters, see Appendix~\ref{app:hparams}. A rank-32 LoRA replication of the d2=100 SFT condition reaches depth-2 pass@64 within the full-fine-tuning seed range (Appendix~\ref{app:lora}).

\paragraph{Models.} We use Qwen3-8B~\citep{Yang2025Qwen3TR} as our primary model and replicate core findings on Qwen3-4B and Qwen3-32B (see Appendix~Table~\ref{tab:4b_replication}) as well as Llama-3.1-8B~\citep{dubey2024llama3} (Appendix~\ref{app:llama}). All training uses full-weight fine-tuning in bfloat16 with FSDP2 and fp32 optimizer states. 

\paragraph{Supervised fine-tuning.} Each SFT run trains on a mixture of 450 depth-1 primitive exemplars unless otherwise noted and varying counts of depth-2 or -3 compositional exemplars. The base configuration across all SFT experiments uses learning rate $5\times 10^{-5}$ with cosine decay, batch size 8, and 3 epochs on 2$\times$H100 GPUs. This recipe was selected from a hyperparameter sweep over learning rate, batch size, and epoch count. Thinking is disabled during both training and inference.

\paragraph{GRPO.} Starting from the best SFT checkpoint, we train with GRPO~\citep{Shao2024DeepSeekMathPT, DeepSeekAI2025DeepSeekR1IR} on 2,500 tasks (500 per depth for depths 1--5). For each prompt, we sample a group of $n=64$ rollouts and compute binary correctness reward (1 if the generated program produces correct output on all held-out input-output pairs and 0 otherwise). Training uses learning rate $2\times 10^{-6}$, KL penalty coefficient 0.005, temperature 0.8, and clip ratio 0.2, on 8$\times$H100 GPUs with VeRL~\citep{sheng2024hybridflow} and vLLM~\citep{kwon2023efficient} for rollout generation. 

\paragraph{PPO.} As a comparison, we also train with PPO~\citep{Schulman2017ProximalPO} from the same SFT checkpoint with GAE~\citep{Schulman2015HighDimensionalCC} ($\gamma=1.0$, $\lambda=0.95$) with a learned value head, matching all other hyperparameters to the GRPO configuration.

\paragraph{Statistical analysis.} Unless stated otherwise, all reported confidence intervals are paired 95\% bootstrap CIs over $n{=}500$ tasks, comparing per-task pass@$k$ indicators across conditions evaluated on identical task sets. Reported $p$-values are from McNemar's exact test on the discordant 2$\times$2 cell. Wilson score intervals are used for marginal proportions. Multi-seed comparisons use paired $t$-tests across matched seeds.

\section{Results}

We organize our findings around four questions: (1) what supervision enables compositional generalization, (2) how it scales with data, (3) whether RL extends it, and (4) whether it transfers across depths.

\subsection{Irreducible supervision is necessary for compositional generalization}
\label{sec:res-irreducible}

\begin{figure*}[h]
    \centering
    \includegraphics[width=0.8\textwidth]{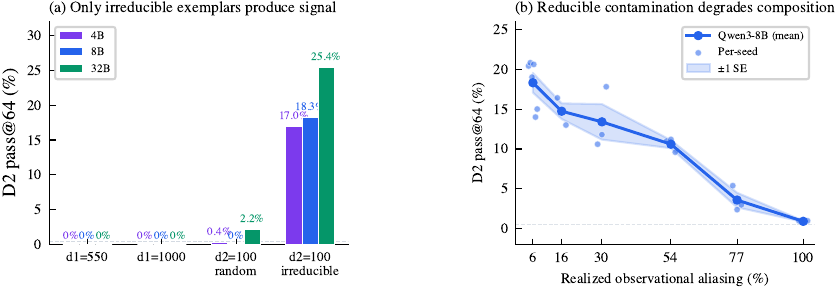}
    \caption{Compositional generalization requires structurally irreducible supervision. (a) Only irreducible compositional exemplars produce signal: d1-only and random-d2 yield 0--2.2\% depth-2 pass@64 across model sizes; 100 irreducible d2 exemplars produce compositional generalization (17.0\% / 18.3 $\pm$3.0\% / 25.4\% for 4B/8B/32B). 8B multi-seed (6/7 seeds, val\_loss criterion); 4B/32B single-seed. (b) Reducible exemplars degrade compositional generalization monotonically: from 100 irreducible depth-2 exemplars, we progressively replace with reducible alternatives (total fixed at 100; 3 seeds per dose; x-axis is classifier-measured realized aliasing). D2 pass@64 declines smoothly from 18.3\% to the $\sim$1\% noise floor with low seed variance (std $\leq$3.9pp), reaching the floor only at high contamination.}
    \label{fig:irred-reducibility}
\end{figure*}

Compositional generalization at depth 2 requires both structural and observational irreducibility. Figure~\ref{fig:irred-reducibility}a and Table~\ref{tab:pipeline-ablation} compare four SFT conditions crossing program structure (irreducible vs.\ reducible) with I/O selection (adversarial vs.\ naive), holding total exemplar count fixed at 100 depth-2 plus 450 depth-1 primitives. The \emph{conjunction}---structurally and observationally irreducible data---reaches 18.3 $\pm$ 3.0\% pass@64 across 6 of 7 seeds passing a val\_loss convergence criterion (7-seed raw mean 16.4 $\pm$ 5.8\%; Appendix~\ref{app:seed-variance}).\footnote{Marginal-contribution comparisons use the 7-seed raw cell-1 mean (16.4\%); the 18.3 $\pm$ 3.0\% headline excludes one val\_loss outlier (val\_loss $>$ 0.4 indicating training failure).} Structurally irreducible programs with naive I/O reach 6.9 $\pm$ 2.7\% (partial signal, well below the conjunction); structurally reducible programs produce $\leq$1\% pass@64 regardless of I/O selection (0.6\% adversarial, 0.1\% naive). Adversarial I/O selection adds 9.5pp to structurally irreducible programs but only 0.5pp to reducible programs; it does not produce compositional signal where none exists.

\begin{table}[h]
\centering
\caption{2$\times$2 ablation crossing program structure with I/O selection at d1=450, d2=100 (Qwen3-8B, $n$=500). Cell 1 multi-seed across 6/7 seeds (val\_loss criterion); cell 2 across 7 seeds, cells 3 and 4 across 3 seeds. See Appendix~\ref{app:seed-variance}.}
\label{tab:pipeline-ablation}
\begin{tabular}{lcc}
\toprule
 & Adversarial I/O & Naive I/O \\
\midrule
Irreducible programs & \textbf{18.3 $\pm$ 3.0} & 6.9 $\pm$ 2.7 \\
Reducible programs   & 0.6 $\pm$ 0.5          & 0.1 $\pm$ 0.2 \\
\bottomrule
\end{tabular}
\end{table}

A direct contamination experiment shows the signal degrades smoothly with reducibility (Figure~\ref{fig:irred-reducibility}b). Holding total exemplar count fixed at 100, we progressively replace irreducible exemplars with reducible alternatives. Depth-2 pass@64 declines monotonically with the realized observational aliasing of the training set---from 18.3\% (clean) through 14.7\% and 13.4\% at 10\% and 25\% nominal contamination to the $\sim$1\% noise floor at full contamination---with low seed-to-seed variance (std $\leq$3.9pp across 3 seeds). Replicating the 10\% and 25\% doses with three additional seeds gives 6-seed means of 15.1\% and 11.4\%, and no seed in any condition reaches the floor until contamination is high. Even the nominally-clean arm carries 6\% residual aliasing, so the cleanest training set is not perfectly irreducible. The degradation is gradual rather than catastrophic: reducible exemplars dilute the compositional signal in proportion to the shortcut structure they introduce. Full results in Appendix~Table~\ref{tab:reducibility_table}.

\subsection{Compositional generalization scales with supervision}
\label{sec:res-comp-gen-scaling}

\begin{figure}[h]
    \centering
    \includegraphics[width=0.5\linewidth]{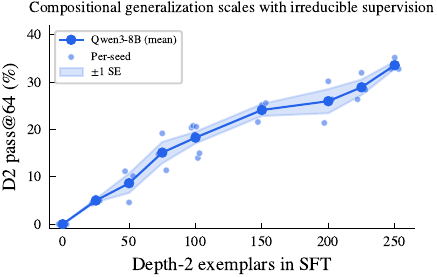}
    \caption{Compositional generalization scales with irreducible supervision. D2 pass@64 vs. d2 exemplar count (Qwen3-8B, 3 seeds per condition, $n$=500). Mean rises 0\% to 33.6\%; $\pm$1 SE shaded. Per-seed variance is higher at d2 $\leq$100 (std up to 9.4pp; Appendix~\ref{app:seed-variance}).}
    \label{fig:comp-gen}
\end{figure}

Section~\ref{sec:res-irreducible} establishes that a 100-exemplar SFT checkpoint produces 18.3 $\pm$ 3.0\% depth-2 pass@64, but leaves open whether this is near saturation or a point on a scaling curve. Figure~\ref{fig:comp-gen} shows depth-2 pass@64 as a function of exemplar count from 0 to 250, replicated with 3 seeds per condition. The mean curve rises smoothly and monotonically from 0\% to 33.6\%; per-seed variance is elevated at d2$\leq$100 (std up to 9.4pp) and lower at d2$\geq$150 (std $\leq$4.4pp), suggesting d2$\geq$150 is a more reliable operating point than d2=100 (Appendix~\ref{app:seed-variance}). Success is distributed across 139 of 427 distinct primitive pairs, with success varying across outer primitives: canvas, hupscale, and downscale are solved at 69\%, 68\%, and 60\% of evaluation tasks, whereas subgrid and crop remain at 0\% (Appendix~Table~\ref{tab:primitive-breakdown}), consistent with partial but genuine compositional generalization rather than pair memorization.

We replicate the core result with Qwen3-4B and Qwen3-32B (Appendix~\ref{app:scale}): the same qualitative pattern holds across all three scales, with absolute performance increasing monotonically with model size (17.0\% / 18.3 $\pm$ 3.0\% / 25.4\% pass@64 at d2=100, and 29.6\% / 33.6 $\pm$ 1.4\% / 36.6\% at d2=250, for 4B / 8B / 32B; 8B is multi-seed, 4B and 32B single-seed), confirming the finding is not scale-specific.

\subsection{RL maintains but does not extend compositional structure}
\label{sec:res-RL}

We first ask whether RL can discover compositional solutions that SFT did not install. It cannot: from a depth-1-only SFT checkpoint, GRPO across four configurations produces zero compositional generalization (depth-2 pass@64 $\leq$1\%; Appendix Table~\ref{tab:rl-sweep-d1only}), and rejection sampling at $k$=512 likewise yields zero correct programs across 500 depth-2 tasks.

\begin{figure}[h]
    \centering
    \includegraphics[width=0.5\linewidth]{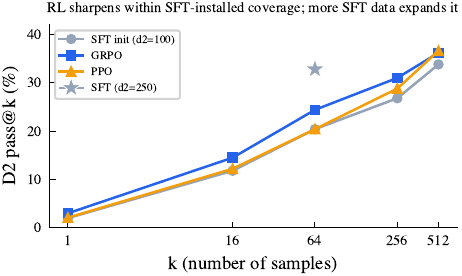}
    \caption{RL sharpens but does not expand compositional coverage. D2 pass@$k$ for SFT init (d2=100), best GRPO, and best PPO across $k=1$ to 512; the gap closes at $k=512$. Star: 2.5$\times$ more SFT data at $k=64$ matches RL at $k=512$.}
    \label{fig:rl-null}
\end{figure}

From a compositional SFT initialization (d1=450, d2=100; depth-2 pass@64 = 20.4\%), GRPO does help at intermediate $k$ (+2.1pp at $k$=64, paired 95\% CI [+1.0, +3.1]pp, unbiased estimator over 512 samples), but the advantage shrinks as $k$ grows and is statistically null by $k$=512 (Figure~\ref{fig:rl-null}; Appendix~\ref{app:add-rl}). RL concentrates probability on solutions the SFT model can already produce; it does not add new ones. The cross-family replication on Llama-3.1-8B base reproduces this pattern (D2 +2.75pp at $k$=128 on a fresh independent draw, GRPO 25.4\% vs.\ SFT 22.65\%; D3/D4 within noise; Appendix~\ref{app:llama}).

Even this sharpening effect is bounded by the quality of the SFT initialization. From a stronger initialization (d2=250, depth-2 pass@64 = 35.2\%), GRPO produces no statistically detectable improvement at any $k$ (peak +1.2pp at $k{=}64$, [$-$1.6, +4.2]pp, $p{=}0.49$; Appendix~\ref{app:cot-sft}), and across four tested initializations, GRPO's marginal lift shrinks monotonically as the initialization improves. No SFT initialization in either model family shows a detectable GRPO improvement on depth 3 or 4. A sweep over learning rate, KL penalty, and depth curriculum confirms these conclusions are not artifacts of a single hyperparameter choice (Appendix~Table~\ref{tab:rl-sweep}), and initializing from SFT with chain-of-thought (CoT) supervision does not change the picture (Appendix~\ref{app:cot-sft}).

The null is not specific to binary outcome reward. Two additional GRPO runs with denser rewards (a verifier-based partial credit for correct sub-expressions, and partial credit for matched held-out I/O pairs) changed rollout behavior but produced no detectable gain in depth-2 or depth-3 pass@64, and GRPO applied only to depth-3 tasks from a depth-2 SFT checkpoint left depth-3 pass@64 at its initialization value (Appendix~\ref{app:rebuttal-rl}). Learned process reward models remain untested.

A log-probability analysis points away from even the classical sharpening mechanism. The SFT initialization already assigns higher probability to correct than incorrect compositions (per-token logprob gap +0.225 nats; Appendix Table~\ref{tab:logprob-rl}). A sharpening account predicts GRPO widens this gap; instead, GRPO narrows it, raising logprob on incorrect programs 3$\times$ faster than on correct ones (Appendix~Table~\ref{tab:logprob-rl}). In terms of efficiency, compositional capability tracks SFT data, not RL optimization: 2.5$\times$ more irreducible SFT data evaluated at $k$=64 matches RL with 8$\times$ more rollouts evaluated at $k$=512.

\subsection{Compositional transfer requires per-depth supervision}
\label{sec:res-comp-cascade}

We test whether compositional ability transfers across depths and find bidirectional but asymmetric transfer. Depth-3 supervision at 250 exemplars produces 6.0\% depth-2 pass@64 (downward) while depth-2 supervision at 250 exemplars yields only 2.2\% depth-3 pass@64 (upward), consistent with depth-3 programs containing depth-2 subtrees by construction. Depth-3 supervision without lower-depth foundation yields near-zero performance (0.2\%), compared to 3.2\% when mixed with d1=450, d2=100, indicating compositional learning is hierarchically dependent on lower-depth skills. Full results in Appendix~Table~\ref{tab:transfer}.

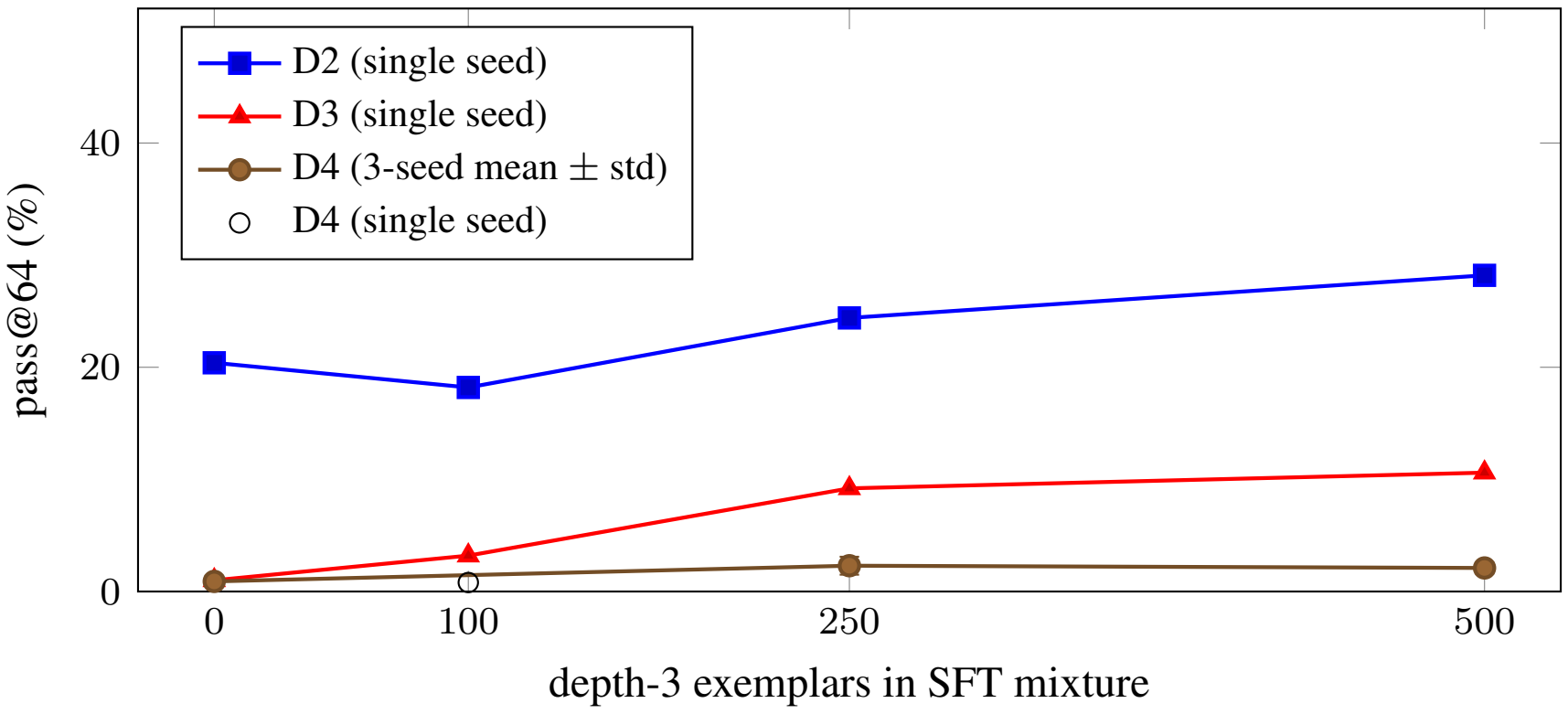
\begin{figure}[h]
\centering
\begin{tikzpicture}
\begin{axis}[
  width=0.85\linewidth, height=5.8cm,
  xlabel={depth-3 exemplars in SFT mixture}, ylabel={pass@64 (\%)},
  xtick={0,100,250,500}, xmin=-30, xmax=530, ymin=0, ymax=52,
  legend pos=north west, legend cell align=left, legend style={font=\footnotesize},
  tick label style={font=\footnotesize}, label style={font=\small}]
\addplot+[mark=square*, thick] coordinates {(0,20.4) (100,18.2) (250,24.4) (500,28.2)};
\addlegendentry{D2 (single seed)}
\addplot+[mark=triangle*, thick] coordinates {(0,1.0) (100,3.2) (250,9.2) (500,10.6)};
\addlegendentry{D3 (single seed)}
\addplot+[mark=*, thick, error bars/.cd, y dir=both, y explicit] coordinates {(0,0.9) +- (0,0.5) (250,2.3) +- (0,0.8) (500,2.1) +- (0,0.5)};
\addlegendentry{D4 (3-seed mean $\pm$ std)}
\addplot+[only marks, mark=o] coordinates {(100,0.8)};
\addlegendentry{D4 (single seed)}
\end{axis}
\end{tikzpicture}
\caption{Depth-$d$ pass@64 (\%) as depth-3 supervision grows (d1=450, d2=100 fixed, $n{=}500$ tasks per depth, paired across conditions). D2 and D3 are single-seed. D4 at d3$\in\{0,250,500\}$ is the mean over 3 paired seeds; error bars show one standard deviation across the 3 seeds. D4 at d3=100 is single-seed. Numeric values and per-seed results are in Appendix Table~\ref{tab:cascade} and Appendix~\ref{app:seed-variance}.}
\label{fig:cascade-main}
\end{figure}

To see whether this transfer scales with additional depth-3 supervision and propagates beyond the trained range, we vary the number of depth-3 exemplars from 0 to 500 in the SFT training mixture (with d1=450, d2=100 fixed) and find depth-dependent effects on pass@64 (Figure~\ref{fig:cascade-main}). Depth-3 generalization scales through roughly 250 exemplars; the d3=500 condition shows no detectable additional gain ([$-$0.8, +3.6], $p$=0.28). Lower-depth performance also benefits (D2 reaches 28.2\% at d3=500), consistent with the downward-transfer pattern noted above.


Depth-4 transfer, although weak, saturates by d3=250. We replicated d3$\in\{$0, 250, 500$\}$ across 3 paired seeds. From a d3=0 baseline of 0.9\% D4 pass@64, d3=250 reaches 2.3 $\pm$ 0.8\% (paired $t$, $n{=}3$ matched seeds, $+$1.33pp, $p{=}0.017$); d3=500 yields no further gain (2.1 $\pm$ 0.5\%; $-$0.13pp, $p{=}0.64$). Compositional capability propagates weakly by one depth and then plateaus. Combined with the Section~\ref{sec:res-RL} result that RL does not extend the SFT-installed ceiling, the picture is that robust performance at depth $d$ requires explicit supervision at depth $d$: neither additional rollouts nor adjacent-depth supervision reliably unlock depth-$(d{+}2)$. Full results in Appendix~Table~\ref{tab:cascade}; per-seed values in Appendix~\ref{app:seed-variance}.


\section{Related work}


Recent work in post-training presents two opposing accounts of what RL adds on top of SFT. The sharpening view holds that RL cannot expand the base policy's reachable set: \citet{Yue2025DoesRL} show that RL gains at small $k$ vanish at large $k$, \citet{Zhao2025EchoCR} show convergence toward dominant pretraining patterns, and \citet{Foster2025IsAG} prove a coverage requirement for efficient RL post-training. The creation view argues the opposite, with scope conditions: RL extends compositional capability when the model has acquired atomic skills via SFT or pretraining~\citep{Yuan2025FromFA, Cheng2025FromAT, Liu2025ProRLPR}.

\citet{Zhang2025Interplay} argue these views are not in fundamental conflict. They show that RL produces capability gains only when pre-training leaves headroom and RL data targets tasks that are difficult but not unreachable. Under their framework, the sharpening view holds in-distribution and the creation view holds when RL data is well-calibrated out-of-distribution. They further show that contextual generalization from RL requires minimal pre-training exposure to the target context, though they do not examine SFT's role. 

The structural property we formalize has precedent in adjacent literature. \citet{Min2019CompositionalQD} showed that many HotpotQA ``multi-hop" questions admit single-passage solutions, and \citet{Trivedi2021MM} filter such shortcuts at the dataset level. \citet{Chen2024BUTTON} similarly construct tool-use training data designed to require each intermediate call, implicitly relying on structural irreducibility. 

\section{Discussion}
\label{sec:discussion}

\subsection{Broader implications}

The standard post-training pipeline (SFT warm-start followed by RL) implicitly assumes that RL will discover compositional strategies from sparse signal. Our results suggest this assumption fails. RL maintains compositional ability but does not create it; the bottleneck is upstream, in whether the SFT data contains examples that require the target composition depth rather than admitting shallower solutions. This gives a practical reading of~\citet{Foster2025IsAG}'s coverage theorem. Empirically, the SFT data structure (irreducibility + adversarial I/O) determines whether compositional solutions get coverage in the policy's reachable set, and RL cannot recover coverage that wasn't installed during SFT.

Our null on RL extension contrasts with two lines of prior work reporting positive compositional gains from post-training. The creation view reports RL gains on compositional tasks given proper SFT initialization~\citep{Yuan2025FromFA, Cheng2025FromAT, Liu2025ProRLPR}, and SKILL-MIX-style experiments report depth transfer when fine-tuning on combined-skill exemplars~\citep{Yu2023SkillMixAF}. Both involve domains with substantial pretraining exposure to the target skills (natural-language reasoning, language-skill composition) where post-training plausibly elicits latent capability supported by pretraining priors. Our setting controls for this: the base model has zero capability on \texttt{arc-dsl}, isolating what post-training itself installs. We do not claim to falsify the positive results from these lines of work; we identify a regime where the SFT-RL pipeline behaves differently because pretraining cannot contribute to the compositional skill being learned.

This has direct implications for data curation. Practitioners building post-training pipelines for compositional tasks (multi-step tool use, multi-hop retrieval, chained function calling) should prioritize verifying that training examples genuinely require the target composition depth. In multi-step tool use, an example is reducible if chained tool calls could be replaced by a single API call. In multi-hop QA, a question is reducible if it can be answered from a single passage~\citep{Min2019CompositionalQD}. Compositional data-construction pipelines for multi-turn function calling already rely on this intuition empirically~\citep{Chen2024BUTTON}; our results predict that training on shortcut-admitting examples provides little compositional signal.

\subsection{Transfer to natural-language composition}
\label{sec:disc-transfer}

If the irreducibility principle generalizes beyond the DSL, then nominally multi-hop but reducible training data should provide no compositional advantage over explicitly single-hop data, even in natural language. We evaluate this on HotpotQA~\citep{Yang2018HotpotQAAD}, classifying bridge questions as reducible or irreducible based on whether the answer can be resolved from a single passage, the natural-language analog of structural irreducibility (Appendix~\ref{app:hotpotqa}).

On held-out irreducible evaluation, SFT on 500 irreducible multi-hop questions yields 0.791 F1, a +3.9pp gain over reducible multi-hop training (0.752) and +5.8pp over single-hop SQuAD training (0.733;~\citep{Rajpurkar2016SQuAD1Q}). Reducible multi-hop and single-hop training differ by just 1.9pp on irreducible eval, within seed variance, suggesting that multi-hop labels alone provide no compositional advantage when the questions admit single-passage shortcuts. All conditions converge on reducible eval ($\approx$0.88 F1), so the gap is specific to genuine multi-hop generalization rather than general QA ability. 

As expected, the effect size is less dramatic compared to the DSL setting. Natural-language QA is less compositionally pure than program synthesis, and parametric knowledge narrows the gap between reducible and irreducible items. The reducibility classification is also lexical and approximate, unlike the mechanically verified irreducibility in the DSL (see Appendix~\ref{app:hotpotqa} for details). We present this as directional evidence that the irreducibility principle extends beyond synthetic domains, not as a standalone finding. Evaluations in domains with more precise reducibility control, such as chained tool use where irreducibility can be verified by ablating intermediate calls, would strengthen this conclusion.

\subsection{Limitations}

\paragraph{Synthetic domain.} Our core experiments use a constructed DSL, providing controlled depth and verifiable irreducibility. The HotpotQA transfer experiment (Section~\ref{sec:disc-transfer}) provides directional evidence that the irreducibility principle extends to natural language, but with smaller effect sizes and approximate reducibility classification. Whether the RL null result generalizes beyond sparse-reward synthesis tasks remains open.

\paragraph{Model scale.} We study 4B, 8B, and 32B parameter models. The widening performance gap at higher exemplar counts (Appendix~Table~\ref{tab:4b_replication}) suggest that scale interacts with supervision efficiency; whether models at 70B+ parameters can extract compositional signal from reducible data remains untested. 

\paragraph{RL algorithms and reasoning regime.} We evaluate GRPO and PPO with outcome reward on single-turn rollouts from the SFT policy, sweeping temperature, learning rate, and KL penalty (Appendix~\ref{app:hparams}). We did not test alternative RL approaches such as process supervision~\citep{Lightman2023LetsVS}, exploration bonuses~\citep{Qu2026POPELT}, or other variants. We synthetically generated CoT (Appendix~\ref{app:cot-grpo}) but natural language reasoning with explicit exploration and backtracking is untested. Our null result characterizes standard RL in the direct-prediction regime, not all possible configurations. For details, see Appendix~\ref{app:hparams}.

\subsection{Future work}

Process reward models that evaluate intermediate subexpressions could provide denser signal than binary outcome reward, potentially overcoming the RL null result in Section~\ref{sec:res-RL}. Similarly, an RL curriculum that gradually increases task depth could also provide a more tractable reward landscape than our uniform-depth setup. The attenuation of generalization across supervisions (Section~\ref{sec:res-comp-cascade}) also motivates depth-progressive training, where a depth-$d$ model generates and filters depth-$d{+}1$ candidates for the next round of SFT, bootstrapping supervision at depths where manual curation is expensive. 

\section{Conclusion}


We study compositional skill emergence during post-training in a controlled program synthesis domain where compositional depth is well-defined and irreducibility is verifiable. Our experiments show that compositional generalization requires irreducibility at both structural and observational levels. Given such supervision, generalization scales smoothly with exemplar count. RL post-training (GRPO and PPO) with outcome reward maintains but does not extend compositional ability beyond the SFT initialization, and the marginal effect of RL itself diminishes as the SFT initialization quality increases. Compositional transfer is bounded: explicit depth-$d$ supervision is required for robust depth-$d$ performance, with weak propagation to one adjacent depth and saturation thereafter. Taken together, the structural properties of SFT data, not the optimization algorithm or data scale, determine compositional capability in this setting.

\section*{Acknowledgements and Disclosures}
Compute for this work was provided by Modal Labs credits that Metis received through the Y Combinator--Modal startup program. Metis was later acquired by DoorDash, and the work was carried out after the acquisition. E.M. is employed by DoorDash. N.A. is affiliated with the Kempner Institute at Harvard University. This work has been made possible in part by a gift from the Chan Zuckerberg Initiative Foundation to establish the Kempner Institute for the Study of Natural and Artificial Intelligence. The authors declare no other third-party support or financial relationships that could be perceived as influencing this work.

\bibliography{references}
\bibliographystyle{abbrvnat}

\newpage
\appendix

\section{DSL specification and task generation}
\label{app:dsl}

\subsection{Primitive catalog}
\label{app:dsl-primitives}

The full \texttt{arc-dsl} contains 160 typed primitives. We include 119 after the following exclusions:

\begin{itemize}
    \item \textbf{Higher-order primitives} (21 excluded): primitives requiring \texttt{Callable} arguments (e.g. \texttt{apply}, \texttt{compose}, \texttt{fork}); these require function-valued intermediate expressions that our typed PCFG cannot currently sample.
    \item \textbf{Type-unsafe and degenerate primitives} (20 excluded): type-unsafe primitives such as  \texttt{first} and \texttt{last} (return \texttt{Any}), \texttt{combine} (pairs incompatible containers), and \texttt{index} (returns \texttt{None} if out of bounds), which produce runtime type errors that corrupt grid cells, and degenerate primitives such as \texttt{identity}, \texttt{repeat}, and \texttt{product} that do not contribute meaningful computation.
\end{itemize}

Included primitives span grid transforms (rotations, mirrors, crops, resizing), color modification (recolor, fill, switch, replace), object detection and filtering (objects, colorfilter, sizefilter, partition), spacial operations (shift, upscale, downscale), arithmetic (increment, decrement, halve, double), logic, set operations, and tuple construction. Two primitives with generic types, \texttt{initset} (\texttt{Any}$\to$\texttt{FrozenSet}) and \texttt{merge} (\texttt{Container[Container]}$\to$\texttt{Container}), are included despite genericity as they enable useful object manipulation chains with low type-safety risk.

\subsection{Program sampling}
\label{app:sampling}

Programs are generated by type-directed top-down synthesis from a typed PCFG. Given a target depth $d$ and a target return type (e.g. \texttt{Grid}), the sampler recursively selects a primitive whose return type matches the target, then fills each argument slot by recursing at depth $d-1$. At depth 0, only constants and the input grid $I$ are eligible.

We enforce a minimum branching constraint for depths $\geq2$: at least one non-spine argument must be filled by a non-leaf (i.e. a nested primitive application rather than a constant or $I$). This prevents degenerate linear chains (e.g. $f(g(h(I, c_1), c_2), c_3)$) where all non-$I$ arguments are constants, ensuring that programs exercise meaningful multi-argument composition.

\subsection{Irreducibility enforcement}
\label{app:irreducibility}

A central requirement for our experimental design is that a program labeled as having depth $d$ genuinely requires depth-$d$ composition---in other words, there should not exist a shallower program that computes the same function. We enforce this using seven layers of filtering, combining static analysis at construction time with behavioral checks at verification time, plus an empirical audit of residual contamination. 

\paragraph{Layer 1: Static collapsible pair-blocking.} 

We maintain a catalog of 85 forbidden parent-child primitive pairs that are known to be algebraically reducible:

\begin{itemize}
    \item \textbf{Dihedral group} $D_4$: symmetry group of the square; compositions of rotations and mirrors that reduce to a single operation (e.g., \texttt{dmirror(hmirror($x$))} $=$ \texttt{rot90($x$)}), including cross-products of mirror-rotation compositions.
    \item \textbf{Involutions}: self-compositions that equal identity (mirrors, flips, inversions).
    \item \textbf{Idempotent operations}: self-compositions that equal self (\texttt{normalize}, \texttt{compress}, \texttt{toindices}).
    \item \textbf{Arithmetic inverses}: \texttt{increment} and \texttt{decrement}, \texttt{halve} and \texttt{double}.
    \item \textbf{Scale round-trips}: \texttt{downscale(upscale($x$,$k$),$k$)} $= x$.
    \item \textbf{Color operation chains}: compositions where the outer operation overwrites the inner's effect, e.g.\ \texttt{recolor(recolor($x$, $c_1$), $c_2$)} $\equiv$ \texttt{recolor($x$, $c_2$)}, and analogous pairs for \texttt{fill} and \texttt{switch\_color}.
    \item \textbf{Semantic extractors after recolor}: extracting color-derived properties after a recolor yields a constant independent of the inner operation, e.g.\ \texttt{numcolors(recolor($x$, $c$))} $\equiv$ \texttt{1}.
    \item \textbf{Absorptions}: compositions where the outer primitive absorbs the inner, e.g.\ \texttt{normalize(crop($x$))} $\equiv$ \texttt{normalize($x$)} when crop preserves the bounding box.
    \item \textbf{Half-operation pairs}: all $4 \times 4$ combinations of \texttt{\{tophalf, bottomhalf, lefthalf, righthalf\}} composed with themselves or each other, which either equal the inner operation or produce a quarter-grid extractable by a single crop.
\end{itemize}

\paragraph{Layer 2: $D_4$ group chain detection and arithmetic inverses.}

For chains of length $\geq 3$ involving geometric transforms, pairwise checks are insufficient. The dihedral group $D_4 = \langle r, s\mid r^4=s^2=(sr)^2=e \rangle$ has elements $\{e, r, r^2, r^3, s, sr, sr^2, sr^3\}$ corresponding to $\{\text{id}, \text{rot90},\text{rot180},\text{rot270}, \text{hmirror}, \text{dmirror}, \text{vmirror},\text{cmirror} \}$. Any composition of $k$ group elements reduces to a single element, so a chain of $k$ geometric transforms is always equivalent to a depth-1 program (e.g. \texttt{rot90(rot90(rot90(rot90($x$))))} $= x$ but no adjacent pair is forbidden). We precompute the $8\times 8$ Cayley table and check whether any contiguous subsequence of ancestor primitives along the spine reduces to the identity or a single non-trivial element already present at a shallower depth. This layer also detects arithmetic inverse chains across non-adjacent ancestors (e.g. \texttt{increment(\ldots decrement(\ldots))}).

\paragraph{Layer 3: Subtree equivalence filter.} 

For each candidate, we extract all shallower sub-expressions from the abstract syntax tree (AST) and test whether any produces the same outputs as the full program on 8 random test grids. This catches cases where an outer operation is a semantic no-op at runtime (e.g., \texttt{fill(f(g($I$)), value, $\emptyset$)} $\equiv$ \texttt{f(g($I$)} when the patch is empty). The filter also checks for identity equivalence (output $\equiv$ input).

\paragraph{Layer 4: Behavior fingerprint cache.}

All depth-1 programs are enumerated for the fingerprint cache: 572 single-grid-argument programs (used by Layer 3 subtree-equivalence checks and by the Layer 7 audit), plus 15 multi-grid-argument programs such as \texttt{hconcat(I,I)} and \texttt{vconcat(I,I)} that require special handling since no grid constants exist as second arguments, for 587 total in the cache. For depths 2--4, large pools are sampled and fingerprinted: 33,000 depth-2, 17,000 depth-3, and 17,000 depth-4 programs. All programs are executed on a fixed set of 29 diagnostic grids chosen to maximize discriminative power (diverse dimensions from $1{\times}1$ to $8{\times}8$, varying color counts, structured and random patterns), and their output tuples are stored as hash keys. Each candidate depth-$d$ program for $d \geq 2$ is executed on the same diagnostic grids and checked against the cache for all depths ${<}d$ in $O(1)$ time. 
 
The size of the fingerprint cache matters for deeper depths. Our initial cache (5{,}000 depth-2 fingerprints) provided ${\approx}$49\% coverage of the type-valid depth-2 space, leaving ${\approx}$10\% of depth-3 training programs reducible---a contamination rate we discovered when the random-d2 ablation (Section~\ref{sec:res-irreducible}) showed unexpectedly weak results. Expanding to 33{,}000 depth-2 fingerprints largely closed this gap; residual aliasing is bounded below 1.5\% (95\% Wilson upper, n=500). All training data used in final results was generated against the expanded cache.

\paragraph{Layer 5: Adversarial prompt grid selection.} 

After a program passes all the above filters, we select the 3 visible input-output pairs for the prompt via a greedy set-cover protocol. The target program is executed on all 29 diagnostic grids. For each grid, we count how many depth-1 confusers it eliminates (a confuser is eliminated when its output on that grid differs from the target's). The grid eliminating the most confusers is selected first, and the process repeats for the remaining 2 prompt slots. 

\paragraph{Layer 6: diagnostic grid hardening of held-out pairs.}

For evaluation data, each task's held-out set is augmented with 29 diagnostic grid I/O pairs. Combined with the exhaustive depth-1 enumeration in Layer 4, this makes aliasing to any \emph{depth-1} program structurally impossible on the evaluation set: any depth-$d$ target whose canonical-grid fingerprint matches a depth-1 program would have been rejected at generation, so no depth-1 confuser can pass the held-out reward check. For aliases at depths $2 \leq d' < d$, the fingerprint cache is a large but non-exhaustive sample (33{,}000 depth-2, 17{,}000 depth-3 programs); these aliases are statistically rare but not structurally ruled out. In practice this matters most for depth-4 evaluation.

The canonical set itself is near-saturated for depth-1 discrimination: it distinguishes 255 of 258 depth-1 function classes (98.8\%), with 3 classes sharing canonical fingerprints with another depth-1 program. The Layer 7 audit on 1{,}029 random grids bounds residual contamination empirically.

We apply diagnostic grid hardening to all evaluation data but not to training data, where the computational cost of per-example hardening is prohibitive at scale. Training data relies on Layers 1--5 with the expanded fingerprint cache.

\paragraph{Layer 7: Empirical audit of residual contamination.}

We empirically validate the pipeline by auditing 500 randomly-selected depth-2 programs from training data against an expanded grid pool of 1{,}029 grids (29 canonical plus 1{,}000 random) and full enumeration of all 572 single-grid-argument depth-1 programs. This is stricter than the 8-grid check used during sampling (Layer 3). We find 2 residual aliases (0.4\%, 95\% Wilson CI [0.1\%, 1.4\%]), both of the form \texttt{underpaint(replace(I, \textsc{five}, X), asobject(I))} $\equiv$ \texttt{replace(I, \textsc{five}, X)}; a semantic redundancy missed by the static pair table (Layer 1) because \texttt{underpaint} is conditionally a no-op depending on the inner program's output. We treat $\approx$1.5\% (the upper bound of the 95\% CI) as a conservative ceiling on training-data contamination. Evaluation data is further hardened by Layer 6, which makes aliasing to any depth-1 program structurally impossible on the evaluation set; aliases at depths $\geq 2$ are bounded by fingerprint-cache coverage rather than ruled out structurally.
 
\section{Training hyperparameters}
\label{app:hparams}

\paragraph{SFT hyperparameter selection.} 

We conducted a sweep over learning rate $\in \{1{\times}10^{-5}, 5{\times}10^{-5}, 1{\times}10^{-4}\}$, batch size $\in \{4, 8, 16\}$, and epochs $\in \{1, 3, 5\}$ on 450 depth-1 exemplars with Qwen3-8B, selecting by depth-1 pass@16 on a held-out validation split. The winner (LR=$5{\times}10^{-5}$, batch=8, 3 epochs) was used for all subsequent SFT experiments without further tuning.

\paragraph{RL hyperparameter selection.} We ran 11 configurations across two algorithms (GRPO, PPO), two model sizes (4B, 8B), four learning rates ($1\times 10^{-6}$ to $5\times 10^{-6}$), three KL coefficients (0, 0.005, 0.01), and two depth configurations (depths 1--5 vs depths 2--5 only). Table~\ref{tab:rl-sweep} summarizes all configurations and outcomes. Common failure modes included catastrophic forgetting at higher learning rates ($5\times 10^{-6}$ on 8B), reward collapse under strong KL penalty (0.01), and reward sparsity when excluding depth-1 tasks. The best configuration (LR=$2\times 10^{-6}$, KL=0.005, temp=0.8, depths 1--5) is reported in the main text alongside PPO, which uses the same hyperparameters with the addition of GAE ($\gamma=1.0$, $\lambda=0.95$) and a learned value head.

\begin{table*}[h]
\centering
\caption{Training hyperparameters for SFT, GRPO, and PPO. All experiments use full-weight fine-tuning, bfloat16 precision, FSDP2 sharding, cosine learning rate schedule with 10\% warmup, gradient clipping at 1.0, and weight decay 0.01.}
\label{tab:hparams}
\begin{tabular}{lcccc}
\toprule
& \textbf{SFT} & \textbf{GRPO} & \textbf{PPO} \\
\midrule
Base model & Qwen3-8B & \multicolumn{2}{c}{Best SFT checkpoint} \\
Learning rate & $5 \times 10^{-5}$ & $2 \times 10^{-6}$ & $2 \times 10^{-6}$ \\
Batch size & 8 & 32 & 32 \\
Epochs & 3 & 5 & 5 \\
Group size ($n$) & --- & 64 & 64 \\
KL coefficient & --- & 0.005 & 0.005 \\
Clip ratio & --- & 0.2 & 0.2 \\
Temperature & --- & 0.8 & 0.8 \\
GAE $\gamma$ / $\lambda$ & --- & --- & 1.0 / 0.95 \\
Max prompt length & 4096 & 1024 & 1024 \\
Max response length & 4096 & 2048 & 2048 \\
Reward & --- & \multicolumn{2}{c}{Binary correctness (all held-out pairs)} \\
Hardware & 2$\times$H100 & 8$\times$H100 & 8$\times$H100 \\
Framework & VeRL (FSDP2) & VeRL + vLLM & VeRL + vLLM \\
\bottomrule
\end{tabular}
\end{table*}

\begin{table*}[h]
\centering
\caption{RL hyperparameter sweep from the compositional SFT checkpoint (d1=450, d2=100 irreducible). All runs use group size 64. The best GRPO configuration (starred) is reported in the main text. The d2-only run is the closest analog to the setup in \citet{Yuan2025FromFA}; in our domain it produces mode collapse rather than compositional discovery, consistent with reward sparsity at depth 2 ($<$2\% rollout success rate). Runs from the depth-1-only SFT checkpoint (no compositional supervision) are reported separately in Table~\ref{tab:rl-sweep-d1only}.}
\label{tab:rl-sweep}
\small
\begin{tabular}{llccccp{4.2cm}}
\toprule
Algo & Model & LR & KL & Temp & Depths & Outcome \\
\midrule
GRPO & 4B & 1e-6 & 0 & 1.0 & mixed & LR too low, pg\_clipfrac=0 \\
GRPO & 4B & 5e-6 & 0 & 0.7 & mixed & D1 improved, D2 false positives \\
GRPO & 8B & 5e-6 & 0 & 0.7 & mixed & Catastrophic forgetting \\
GRPO & 8B & 1e-6 & 0.01 & 0.7 & mixed & KL too high, reward declined \\
GRPO & 8B & 2e-6 & 0.005 & 0.8 & 1--5 & D1 dominates, D2 flat \\
GRPO & 8B & 2e-6 & 0.005 & 0.9 & 2--5 & Reward too sparse \\
GRPO$^\star$ & 8B & 2e-6 & 0.005 & 0.8 & 1--5 & Best: +3.2pp D2 p@16 (n.s.) \\
PPO & 8B & 2e-6 & 0.005 & 0.8 & 1--5 & No signal, within noise \\
GRPO & 8B & 2e-6 & 0.005 & 0.8 & 1--5 & CoT enabled; SFT init non-CoT, model stayed ${\approx}$20 tok; D2 flat \\
GRPO & 8B & 2e-6 & 0.005 & 1.0 & 2 only & Mode collapse: p@1 2\%$\to$11\%, p@64 24\%$\to$18\% (373/3500 steps) \\
GRPO & 8B & 2e-6 & 0.005 & 1.0 & 1--2 &  D2 +1.4pp (n.s.); flat reward after step 75 \\
\bottomrule
\multicolumn{7}{l}{\footnotesize $^\star$Best configuration; reported in main text.}
\end{tabular}
\end{table*}

\begin{table*}[h]
\centering
\caption{RL runs from the primitive-only SFT checkpoint (d1=450, no depth-2 supervision), testing whether GRPO can discover compositional structure without compositional SFT. All runs use group size 64. Evaluation here is pre-dealiasing; after dealiasing, all depth-2 gains reduce to zero (the apparent D2 signal in v2 was entirely attributable to depth-aliased programs). None of these runs produced genuine compositional generalization.}
\label{tab:rl-sweep-d1only}
\small
\begin{tabular}{llccccp{4.2cm}}
\toprule
Algo & Model & LR & KL & Temp & Depths & Outcome \\
\midrule
GRPO & 4B & 1e-6 & 0 & 1.0 & mixed & LR too low, pg\_clipfrac=0 \\
GRPO & 4B & 5e-6 & 0 & 0.7 & mixed & D1 improved (38.7\%$\to$50.5\%); D2 apparent gains all aliased \\
GRPO & 8B & 5e-6 & 0 & 0.7 & mixed & Catastrophic forgetting (D1 collapsed) \\
GRPO & 8B & 1e-6 & 0.01 & 0.7 & mixed & KL too high, reward declined \\
\bottomrule
\end{tabular}
\end{table*}

\paragraph{Compute.} Total compute across all reported and exploratory experiments was approximately 3,800 GPU-hours on a mix of NVIDIA H100 (8$\times$80GB and 2$\times$80GB nodes via Modal) and H200 (8$\times$141GB nodes) hardware. Exploratory and failed runs not described in the main text or appendices, including: hyperparameter sweeps that did not change conclusions, data-generation iterations during pipeline development, recipe-mismatched cross-family runs later superseded by matched-recipe replications, and re-evaluations following bug fixes, account for roughly 18\% of the total.

\paragraph{Assets and licenses.} Base models: Qwen3-\{4B,8B,32B\} (Tongyi Qianwen License) and Llama-3.1-8B / Llama-3.1-8B-Instruct (Llama 3.1 Community License). DSL: Hodel's arc-dsl (MIT). Frameworks: VeRL, vLLM, HuggingFace Transformers (Apache 2.0); PyTorch (BSD-3-Clause). HotpotQA (CC BY-SA 4.0) and SQuAD 2.0 (CC BY-SA 4.0) used in Section~\ref{sec:disc-transfer} and Appendix~\ref{app:hotpotqa}. All assets used consistent with their respective licenses; generated data and trained checkpoints inherit base-model license terms; our code will be released under MIT.



\section{SFT scaling and ablations}
\label{app:add-sft}

\begin{figure*}[h]
    \centering
    \includegraphics[width=\textwidth]{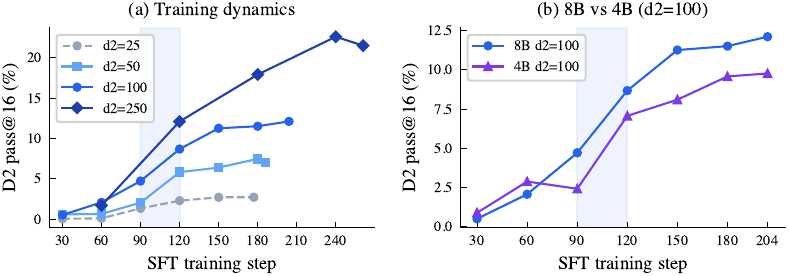}
    \caption{Training dynamics of compositional generalization. (a) Depth-2 pass@16 over SFT training steps for varying numbers of irreducible depth-2 exemplars (Qwen3-8B). All conditions with d2$\geq$50 show compositional signal emerging between steps 90–120 (shaded region). (b) 8B vs 4B comparison at d2=100. The 8B model reaches higher absolute performance but both models show similar emergence timing. Held-out eval $n=500$.
    /depth.}
    \label{fig:supp-train-dynamics}
\end{figure*}

\begin{table*}[h]
\centering
\caption{Best-checkpoint results across supervision conditions (held-out $n{=}500$/depth, $k{=}64$, single-seed). All runs use Qwen3-8B with LR=$5{\times}10^{-5}$, batch size 8, 3 epochs, full-weight fine-tuning. Unless noted, each run uses 450 depth-1 exemplars plus the indicated number of irreducible depth-2 exemplars. For D2 pass@64 across seeds (3 seeds per condition), see Table~\ref{tab:seed-variance}.}
\label{tab:supervision_sweep}
\begin{tabular}{lrrrrr}
\toprule
Config & D2 p@1 & D2 p@16 & D2 p@64 & D3 p@64 & D4 p@64 \\
\midrule
d2=0 (d1=1000$^\dagger$) & 0 & 0 & 0 & 0 & 0 \\
d2=25 & 0.6 & 2.7 & 4.4 & 0.2 & 0.2 \\
d2=50 & 1.3 & 7.0 & 11.2 & 0.4 & 0.6 \\
d2=75 & 1.2 & 7.6 & 14.8 & 1.2 & 1.0 \\
d2=100 & 2.0 & 12.1 & 20.4 & 1.0 & 1.4 \\
d2=150 & 2.5 & 13.9 & 21.6 & 2.0 & 1.0 \\
d2=200 & 2.1 & 12.5 & 21.4 & 2.2 & 1.2 \\
d2=225 & 3.7 & 17.7 & 26.4 & 2.0 & 2.2 \\
d2=250 & 5.5 & 21.5 & 32.8 & 2.0 & 2.2 \\
\bottomrule
\multicolumn{6}{l}{\footnotesize $^\dagger$1000 depth-1 only (no depth-2); size-matched control.}
\end{tabular}
\end{table*}

\begin{table}[h]
\centering
\caption{Ablation controls (unique to this table; the d2=100 irreducible baseline is in Table~\ref{tab:supervision_sweep}, and the d1=1000 size-matched control is the d2=0 row of that table). Random-d2 supervision and depth-1-only data of any size yield 0\% depth-2 pass@64. Qwen3-8B, single-seed, held-out set ($n{=}500$/depth); see Appendix~\ref{app:seed-variance} for variance estimates near the threshold.}
\label{tab:ablation}
\begin{tabular}{lrrr}
\toprule
Condition & D2 p@16 & D2 p@64 & D3 p@64 \\
\midrule
d2=100 random  & 0 & 0 & 0 \\
d1=550 (no d2) & 0 & 0 & 0 \\
d1=650 (no d2) & 0 & 0 & 0 \\
\bottomrule
\end{tabular}
\end{table}

\begin{table*}[h]
\centering
\caption{Contamination with reducible exemplars degrades compositional generalization monotonically. Starting from 100 irreducible depth-2 exemplars, we progressively replace irreducible programs with reducible alternatives (total fixed at 100). The x-axis is the classifier-measured \emph{realized} observational aliasing: nominal replacement fractions of 0/10/25/50/75/100\% correspond to 6/16/30/54/77/100\% realized aliasing, so even the nominally-clean arm carries 6\% residual. Three seeds per condition (program identities fixed across seeds; seeds vary the train/val partition and ordering); the 0\% baseline is the irreducible+adversarial cell of the 2$\times$2 ablation (Section~\ref{sec:res-irreducible}, multi-seed, 6/7 seeds). D2 pass@64 declines monotonically with low seed variance (std $\leq$3.9pp), reaching the $\sim$1\% noise floor only at high contamination ($\geq$77\% realized). Qwen3-8B, held-out eval ($n=500$/depth).}
\label{tab:reducibility_table}
\begin{tabular}{rrrrr}
\toprule
Realized aliasing (\%) & Seed 42 & Seed 43 & Seed 44 & Mean $\pm$ std \\
\midrule
6   & 20.4 & 20.8 & 4.8$^*$ & 18.3 $\pm$ 3.0 \\
16  & 16.4 & 14.8 & 13.0    & 14.7 $\pm$ 1.7 \\
30  & 10.6 & 11.8 & 17.8    & 13.4 $\pm$ 3.9 \\
54  & 11.0 & 11.2 & 9.6     & 10.6 $\pm$ 0.9 \\
77  & 5.4  & 2.4  & 3.0     & 3.6  $\pm$ 1.6 \\
100 & 1.0  & 0.8  & 1.0     & 0.9  $\pm$ 0.1 \\
\bottomrule
\multicolumn{5}{l}{\footnotesize $^*$Seed 44 fails val\_loss $<$ 0.4 convergence criterion at d2=100;} \\
\multicolumn{5}{l}{\footnotesize Mean $\pm$ std at 0\% reducible reflects 6/7 seeds passing the criterion.} \\
\end{tabular}
\end{table*}

\begin{table*}[h]
\centering
\caption{Bidirectional transfer is asymmetric: depth-3 supervision produces some depth-2 generalization (downward), while depth-2 supervision yields weaker depth-3 transfer (upward). D4 dips at small d3 counts (0.8 at d3=100 vs.\ 1.4 at baseline), recovering at d3=250, consistent with distributional interference before sufficient depth-3 signal accumulates. All runs use Qwen3-8B, eval ($n{=}500$/depth), single-seed.}
\label{tab:transfer}
\begin{tabular}{lrrrrr}
\toprule
Config & D2 p@16 & D2 p@64 & D3 p@16 & D3 p@64 & D4 p@64 \\
\midrule
d1=450 + d2=100 (baseline) & 12.1 & 20.4 & 0.7 & 1.0 & 1.4 \\
d2=100 only (no d1) & 3.3 & 7.6 & 0.4 & 0.6 & 0.6 \\
d2=250 only (no d1) & 15.3 & 25.8 & 1.3 & 2.2 & 1.6 \\
d3=100 only & 0.2 & 0.8 & 0.1 & 0.2 & 0 \\
d3=250 only & 2.0 & 6.0 & 0.7 & 2.2 & 0.2 \\
\bottomrule
\end{tabular}
\end{table*}

\begin{table}[h]
\centering
\caption{Explicit depth-3 supervision scales D3 generalization and enables weak D4 emergence. Base: 450 d1 + 100 d2. D2 shows a non-significant dip at small d3 counts (14.8 at d3=50, 18.2 at d3=100; paired CIs include zero) and rises significantly above baseline at d3$\geq$250. All runs use Qwen3-8B, eval ($n{=}500$/depth), single-seed.}
\label{tab:cascade}
\begin{tabular}{rrrr}
\toprule
d3 exemplars & D2 p@64 & D3 p@64 & D4 p@64 \\
\midrule
0 & 20.4 & 1.0 & 1.4 \\
25 & 17.4 & 2.2 & 1.2 \\
50 & 14.8 & 1.6 & 1.0 \\
100 & 18.2 & 3.2 & 0.8 \\
250 & 24.4 & 9.2 & 3.0 \\
500 & 28.2 & 10.6 & 2.4 \\
\bottomrule
\end{tabular}
\end{table}

\subsection{Seed variance across experiments}
\label{app:seed-variance}

\subsubsection{Overview}

Multi-seed analysis identifies a consistent pattern: seed variance is elevated at intermediate-quality SFT supervision and lower at both extremes (insufficient or saturated). We refer to this intermediate region, where supervision is barely sufficient for compositional learning, as the \emph{supervision threshold}. Specifically, we observe high variance at d2=50 and d2=100 in the supervision sweep (std up to 9.4pp across 3 seeds; \S\ref{app:supervision-seeds}) and at the irreducible cells of the 2$\times$2 pipeline ablation at d2=100 (std 5.8pp and 2.74pp respectively across 7 seeds each; \S\ref{app:cell-seeds}). Variance is low at d2$\geq$150 (uniformly successful runs across 3 seeds), at d2$\leq$25 / fully-reducible (uniformly at noise floor), and across the contamination dose-response, where every dose has low seed variance (std $\leq$4.3pp across 3--6 seeds; \S\ref{app:other-seeds}). With $n{=}3$ to $n{=}7$ seeds per cell we cannot quantitatively characterize this variance structure, but the descriptive pattern---supervision-threshold-region sensitivity---provides context for the multi-seed values reported in the main text and tables.

\subsubsection{Supervision sweep three-seed replication}
\label{app:supervision-seeds}

The single-seed supervision sweep (Table~\ref{tab:supervision_sweep}) showed an apparent plateau at d2=100--200 followed by a jump to 32.8\% at d2=250. To test whether this reflected real saturation or seed variance, we replicated each condition with seeds 43 and 44 (the original sweep is seed 42). Table~\ref{tab:seed-variance} reports per-seed values.

\begin{table*}[h]
\centering
\caption{Three-seed replication of the d2 supervision sweep (Qwen3-8B, eval $n{=}500$, $k{=}64$, final checkpoint; SFT recipe LR=$5{\times}10^{-5}$, batch 8, 3 epochs). The d2=50 and d2=100 conditions exhibit stochastic seed failures; d2$\geq$150 achieves zero failures across 3 seeds.}
\label{tab:seed-variance}
\begin{tabular}{rrrrrrr}
\toprule
d2 & seed 42 & seed 43 & seed 44 & mean & std & failures \\
\midrule
0   & 0.0  & 0.0    & 0.0    & 0.0  & 0.0 & 0/3 \\
25  & 4.4  & 5.6    & 5.0    & 5.0  & 0.6 & 0/3 \\
50  & 11.2 & 4.6$^*$ & 10.2  & 8.7  & 3.6 & 1/3 \\
75  & 14.8 & 19.2   & 11.4   & 15.1 & 3.9 & 0/3 \\
100 & 20.4 & 20.8   & 4.4$^*$ & 15.2 & 9.4 & 1/3 \\
150 & 21.6 & 25.2   & 25.6   & 24.1 & 2.2 & 0/3 \\
200 & 21.4 & 30.2   & 26.4   & 26.0 & 4.4 & 0/3 \\
225 & 26.4 & 32.0   & 28.4   & 28.9 & 2.8 & 0/3 \\
250 & 32.8 & 35.2   & 32.8   & 33.6 & 1.4 & 0/3 \\
\bottomrule
\multicolumn{7}{l}{\footnotesize $^*$Stochastic failure: model converges with low train loss to a confident} \\
\multicolumn{7}{l}{\footnotesize non-compositional minimum (logprob on its own outputs is higher than} \\
\multicolumn{7}{l}{\footnotesize successful seeds, but pass@64 is at noise floor).} \\
\end{tabular}
\end{table*}

The mean curve is monotonic across d2 conditions, dispelling the apparent plateau at d2=100--200 visible in the single-seed sweep. Per-seed variance is elevated at d2$\leq$100 (std up to 9.4pp) compared to d2$\geq$150 (std $\leq$4.4pp). Two of 24 runs failed to learn (pass@64 $<$5\%, marked with $^*$), both at d2$\leq$100; in failed runs the model converged with low train loss and confident outputs to a non-compositional minimum (logprob on its own outputs is higher than for successful seeds at the same d2 level). With $n{=}3$ per condition we cannot quantitatively characterize the failure-rate function or the variance structure across d2 levels, but observe that runs at d2$\geq$150 are uniformly successful in our sample, suggesting d2$\geq$150 as a more reliable operating point than d2=100.

\subsubsection{2$\times$2 pipeline ablation: cell-level multi-seed}
\label{app:cell-seeds}

The 2$\times$2 pipeline ablation in Section~\ref{sec:res-irreducible} reports the cell-1 conjunction (irreducible + adversarial) and cell-2 (irreducible + naive I/O) cells multi-seed; cells 3 and 4 (reducible) are 3 seeds each. \textbf{Cell 1 (irreducible + adversarial)}: 7 seeds at d2=100 give raw mean 16.4 $\pm$ 5.8\% pass@64, with 4 seeds at 19--21\%, 2 at 14--15\%, and 1 at 4.8\% (elevated training val\_loss $>$ 0.4, used as a principled convergence exclusion criterion); excluding the val\_loss outlier gives 18.3 $\pm$ 3.0\% across 6 seeds. \textbf{Cell 2 (irreducible + naive I/O)}: 7 seeds at d2=100 give 6.94 $\pm$ 2.74\% pass@64 (range 1.6--10.4\%), substantially above the noise floor of cells 3 and 4 but well below the conjunction. \textbf{Cell 3 (reducible + adversarial)}: 0.6 $\pm$ 0.5\%. \textbf{Cell 4 (reducible + naive)}: 0.13 $\pm$ 0.23\%. Cells 1 and 2 share the elevated variance characteristic of the d2=100 supervision-threshold region (cf.\ supervision sweep std 9.4pp at d2=100); the qualitative 2$\times$2 ordering holds across seeds.

\paragraph{Aliasing-rate analysis.} Per-seed prompt-aliasing rate (the fraction of evaluation prompts whose visible I/O pairs admit a depth-1 explanation) does not predict cell-2 pass@64. Across the 7 cell-2 seeds, aliasing rates range from 5\% to 12\% and pass@64 ranges from 1.6\% to 10.4\%, but the two are uncorrelated. The seed with the lowest aliasing rate (5\%) reached 8.2\% pass@64 (mid-range), while the seed with the highest aliasing rate (12\%) reached 10.4\% pass@64 (the highest). The lowest-pass seed (1.6\%) had 6.2\% aliasing---second-lowest---ruling out an explanation in which more aliased prompts produce more solves. The 6.9\% population mean therefore reflects genuine partial compositional generalization that arises from structural irreducibility in the training programs, not from artificially solvable aliased prompts.

\paragraph{Marginal-contribution analysis.} Both factors contribute substantially, with structural irreducibility carrying the larger marginal effect. Holding I/O selection fixed at adversarial, structural irreducibility raises pass@64 by +15.8pp (0.6\% reducible $\to$ 16.4\% irreducible, using 7-seed raw mean for comparability); holding I/O fixed at naive, the structural effect is +6.8pp (0.13\% $\to$ 6.94\%). Conversely, holding program structure fixed at irreducible, adversarial I/O selection adds +9.5pp (6.94\% $\to$ 16.4\%); on reducible programs, adversarial I/O adds only +0.5pp (0.13\% $\to$ 0.6\%). Adversarial I/O selection's contribution is therefore conditional on structural irreducibility---it amplifies an existing compositional signal but cannot manufacture one.

\subsubsection{Other observations}
\label{app:other-seeds}

Two additional non-monotonicities in single-seed experiments are consistent with threshold-region seed variance, with multi-seed evidence for the contamination experiment.

\paragraph{Contamination dose-response (per-seed).} Per-seed depth-2 pass@64 at each dose is reported in Table~\ref{tab:reducibility_table}; seed variance is low at every dose (std $\leq$3.9pp across 3 seeds), and no seed collapses to the noise floor until contamination is high (e.g.\ at 10\% reducible the three seeds give 16.4\% / 14.8\% / 13.0\%, and at 25\% they give 10.6\% / 11.8\% / 17.8\%). We additionally replicated the 10\% and 25\% doses with three more seeds (45--47): the 6-seed means are 15.1$\pm$2.0\% and 11.4$\pm$4.3\% respectively (one 25\% run, seed 46, reached only step 140 and sits at 4.4\%; all others are $\geq$10.6\%). The dose-response is therefore a smooth monotonic decline---reducible contamination dilutes the compositional signal in proportion to the aliasing it introduces, reaching noise only at high contamination---rather than a seed-dependent collapse.

\paragraph{Depth-transfer D2 dip (single-seed).} In the depth-transfer experiments (Table~\ref{tab:cascade}), D2 dips at small d3 counts (14.8 at d3=50, 18.2 at d3=100) before recovering. The corresponding paired CIs at d3=50 and d3=100 vs d3=0 baseline include zero, also consistent with seed variance rather than systematic interference. We did not run multi-seed replications for the D2 dip due to compute cost; we report the dip here to flag the ambiguity rather than assert a single-seed point estimate as canonical at these intermediate cells.

\subsection{Scale replication across model sizes}
\label{app:scale}

To verify that the irreducibility requirement is not a small-model artifact, we replicate the headline d2=100 condition at Qwen3-4B and Qwen3-32B alongside our primary 8B model. Table~\ref{tab:scale-d2-100} reports D2 pass@64 across all three sizes for the three core conditions. The qualitative three-condition pattern (d1-only $\approx$ random-d2 $\ll$ irreducible-d2) replicates at every scale, and absolute performance on the irreducible condition scales monotonically with model size (17.0\% $\to$ 20.4\% $\to$ 25.4\% from 4B to 32B). The transfer scales weakly with model size: 32B reaches D3/D4 pass@64 of 1.2\%/2.4\% vs 8B's 1.0\%/1.4\% (4B: 1.4\%/1.0\%).


\begin{table}[h]
\centering
\caption{Scale replication at d2=100 and d2=250 (D2 pass@64; final-step checkpoint, held-out eval $n{=}500$/depth, $k{=}64$). All conditions use the standard SFT recipe. 8B irreducible cells are multi-seed (mean $\pm$ std; d2=100 across 6/7 seeds with val\_loss criterion, d2=250 across 3 seeds); other cells are single-seed. Irreducible d2 supervision produces compositional generalization at every scale, while d1-only and random-d2 controls remain at or near the noise floor; the irreducible-d2 effect strengthens with model size at both supervision levels. The full 4B depth-2 supervision sweep is in Table~\ref{tab:4b_replication}.}
\label{tab:scale-d2-100}
\begin{tabular}{lrrr}
\toprule
Condition          & 4B   & 8B   & 32B  \\
\midrule
d1-only            & 0.0  & 0.0  & 0.0  \\
d2=100 random      & 0.4  & 0.0  & 2.2  \\
d2=100 irreducible & \textbf{17.0} & \textbf{18.3 $\pm$ 3.0} & \textbf{25.4} \\
d2=250 irreducible & 29.6 & 33.6 $\pm$ 1.4 & \textbf{36.6} \\
\bottomrule
\end{tabular}
\end{table}

We additionally ran a full d2 supervision sweep at 4B to verify the scaling-curve shape (Table~\ref{tab:4b_replication}); the qualitative pattern matches 8B at lower absolute performance, with the gap widening from 0.4pp at d2=25 to 3.2pp at d2=250, suggesting larger models extract more from the same data budget.

\begin{table}[h]
\centering
\caption{4B replication of the d2 supervision sweep. Qwen3-4B shows the same qualitative pattern as Qwen3-8B at lower absolute performance, with the gap widening at higher supervision counts. Both models use matched SFT hyperparameters (LR=$5{\times}10^{-5}$, batch 8, 3 epochs),  eval ($n{=}500$/depth), single-seed per cell.}
\label{tab:4b_replication}
\begin{tabular}{rrr}
\toprule
d2 exemplars & 8B D2 p@64 & 4B D2 p@64 \\
\midrule
25  & 4.4  & 4.0  \\
50  & 11.2 & 9.6  \\
75  & 14.8 & 13.2 \\
100 & 20.4 & 17.0 \\
250 & 32.8 & 29.6 \\
\bottomrule
\end{tabular}
\end{table}

\section{RL post-training}
\label{app:add-rl}

This appendix collects the per-checkpoint and per-$k$ details supporting the main-text Section~\ref{sec:res-RL} claims: GRPO from a compositional SFT initialization sharpens at intermediate $k$ but does not extend compositional coverage, and the effect attenuates as the SFT init quality increases.

\subsection{Training dynamics and best-checkpoint comparison}
\label{app:rl-trajectories}

\begin{figure}[h]
    \centering
    \includegraphics[width=0.5\linewidth]{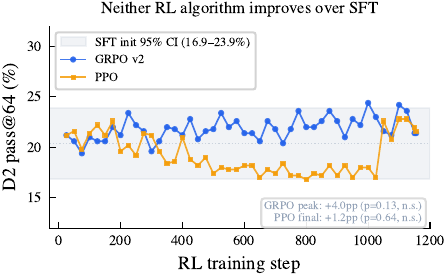}
    \caption{Depth-2 pass@64 over RL training steps for best GRPO and best PPO runs from the d2=100 SFT checkpoint. Both remain within the 95\% binomial CI of the SFT initialization throughout training. Qwen3-8B, 500 held-out irreducible depth-2 tasks.}
    \label{fig:rl-train-dynamics}
\end{figure}

\begin{table*}[h]
\centering
\caption{RL post-training results: best checkpoint per algorithm from the SFT d2=100 checkpoint (eval $n{=}500$). Best step selected by D2 pass@64. D3 remains within noise of the SFT initialization across all checkpoints. Single-seed for each algorithm.}
\label{tab:rl}
\begin{tabular}{lrrrr}
\toprule
Condition & D2 p@16 & D2 p@64 & D3 p@16 & D3 p@64 \\
\midrule
SFT init (d2=100)            & 12.1 & 20.4 & 0.5 & 1.0 \\
GRPO best (step 1000) & 15.3 & 24.4 & 0.6 & 1.6 \\
PPO best (step 100)   & 12.4 & 21.4 & 0.8 & 1.8 \\
\bottomrule
\multicolumn{5}{l}{\footnotesize Full training dynamics including late-step regression in Figure~\ref{fig:rl-train-dynamics};} \\
\multicolumn{5}{l}{\footnotesize per-$k$ paired CIs in main-text Section~\ref{sec:res-RL}.} \\
\end{tabular}
\end{table*}

\subsection{Log-probability dynamics during GRPO}
\label{app:rl-logprob}

The SFT initialization already discriminates correct from incorrect compositions in its logit distribution (per-token logprob gap $+0.225$ nats; see also Appendix~\ref{app:logprob} for the conditional logprob analysis across SFT supervision levels). A classical sharpening account would predict GRPO amplifies this gap by raising logprob on correct programs faster than on incorrect ones. Empirically, the opposite holds: GRPO raises logprob on incorrect programs $3{\times}$ faster than on correct ones (Table~\ref{tab:logprob-rl}), narrowing the discrimination gap while pass@64 barely moves. We attribute this to reward sparsity at depth 2 ($<$2\% rollout success rate): GRPO's group-relative advantage signal cannot isolate the correct mode under such sparse positives, so the distribution sharpens near-uniformly, consistent with the pass@$k$ convergence in Figure~\ref{fig:rl-null}.

\begin{table*}[h]
\centering
\caption{Log-probability dynamics during GRPO training (Qwen3-8B, d2=100 SFT init).}
\label{tab:logprob-rl}
\small
\begin{tabular}{rcccc}
\toprule
GRPO step & D2 p@64 & lp(correct) & lp(incorrect) & Gap (nats) \\
\midrule
25 (init)    & 21.2\% & $-0.127$ & $-0.352$ & $+0.225$ \\
400          & 21.2\% & $-0.116$ & $-0.293$ & $+0.177$ \\
1155 (final) & 21.4\% & $-0.107$ & $-0.286$ & $+0.179$ \\
\bottomrule
\end{tabular}
\end{table*}

\section{Additional RL and LoRA experiments}
\label{app:rebuttal-rl}

This appendix reports experiments run to test whether the RL null result is specific to sparse binary outcome reward, whether depth-3 reward alone can install depth-3 ability after depth-2 SFT, and whether the SFT result depends on full-weight fine-tuning. All runs use Qwen3-8B from the d1=450, d2=100 SFT checkpoint (D2 pass@64 = 20.4\%, D3 pass@64 = 1.0\%) with the GRPO configuration of Appendix~\ref{app:hparams}, unless noted. Evaluation uses the same 500 held-out tasks per depth, and comparisons to the initialization are paired on the same tasks. All GRPO runs here are single-seed.

\subsection{Denser reward in GRPO}
\label{app:dense-reward}

\paragraph{Verifier-based intermediate credit.} Because the task generator retains reference programs, we test a verifier-based analogue of intermediate-step credit. An otherwise incorrect rollout receives reward 0.3 if a proper sub-expression of the generated program is behaviorally equivalent, on every held-out input, to an intermediate of the reference program; complete solutions receive 1.0. We chose 0.3 to give a meaningful learning signal while keeping complete solutions substantially more rewarding, and excluded identity and bare-first-step outputs to reduce trivial reward farming. This isolates the effect of denser intermediate feedback from errors in a learned reward model. It only rewards progress along the reference decomposition and does not substitute for a learned process reward model.

The credited-sub-expression rate on depth-2+ rollouts rose from 18\% to 36\% over training, and mean parsed program depth rose from 1.66 to 1.95. Complete-task performance did not improve at any evaluated checkpoint (Table~\ref{tab:dense-verifier}). Final-versus-initialization paired comparisons were not significant: depth-2 pass@64 changed by $+$0.6pp (95\% CI [$-$2.4, $+$3.6]pp; $p{=}0.79$) and depth-3 by $+$0.8pp (95\% CI [0.0, $+$1.8]pp; $p{=}0.22$). The shaping reward increased sampling of the credited intermediate behavior without a detectable increase in full task completion under the tested rollout budget.

\begin{table}[h]
\centering
\caption{GRPO with verifier-based intermediate credit (single seed). Pass@64 (\%) on 500 held-out tasks per depth.}
\label{tab:dense-verifier}
\small
\begin{tabular}{lcc}
\toprule
Step & D2 pass@64 & D3 pass@64 \\
\midrule
SFT init & 20.4 & 1.0 \\
100 & 21.2 & 1.6 \\
200 & 19.8 & 1.8 \\
300 (final) & 21.0 & 1.8 \\
\bottomrule
\end{tabular}
\end{table}

\paragraph{Dense final-output credit.} As a complementary check, failed rollouts earned 0.3 times the fraction of held-out I/O pairs matched. This reward also changed rollout behavior: among non-solving rollouts on depth-2+ tasks, mean response depth rose from 1.21 to 2.01 and the share of shallow (depth-1) attempts fell from 79\% to 0\%. We observed no evidence of reward hacking in audited rollouts (the average partial reward among failed rollouts decreased from 0.113 to 0.055). Performance stayed numerically close to binary-reward GRPO at every matched checkpoint (final D2 pass@64 20.2\% vs.\ 21.0\%).

Together, neither denser reward produced a detectable gain, so the null is not solely an artifact of binary reward sparsity. Learned process reward models and broader process supervision remain untested.

\subsection{Depth-3 reward after depth-2 SFT}
\label{app:d3-rl}

Starting from the d1=450, d2=100 checkpoint, we trained GRPO exclusively on depth-3 tasks for 150 steps with the hyperparameters of the main GRPO configuration. This tests whether policy-gradient feedback on depth-3 tasks can substitute for depth-3 demonstrations after lower-depth SFT.

\begin{table}[h]
\centering
\caption{GRPO on depth-3 tasks only, from the d1=450, d2=100 SFT checkpoint (single seed).}
\label{tab:d3-rl}
\small
\begin{tabular}{lcc}
\toprule
Step & D2 pass@64 & D3 pass@64 \\
\midrule
SFT init & 20.4 & 1.0 \\
150 (final) & 21.2 & 1.2 \\
\bottomrule
\end{tabular}
\end{table}

Successful rollouts were rare throughout (0--1 per 2048), and almost all GRPO groups had no reward variation and supplied no gradient signal. Depth-3 pass@64 stayed at the initialization value (paired CI [$-$0.6, $+$1.2]pp, $p{=}1.0$) while depth-2 performance was preserved. By comparison, 100 and 250 direct depth-3 SFT exemplars reach 3.2\% and 9.2\% depth-3 pass@64 (Table~\ref{tab:cascade}). This supports a coverage-limited interpretation under the tested rollout budget; it does not establish that prolonged or exploration-enhanced RL could never succeed. The null is also not solely due to absent reward: in the denser-reward runs the policy received nonzero verifier-based intermediate credit on depth-3 tasks throughout, and depth-3 pass@64 likewise did not improve.

\subsection{LoRA replication of SFT}
\label{app:lora}

We repeated the d2=100 SFT condition with rank-32 LoRA and an otherwise identical recipe, at two learning rates (Table~\ref{tab:lora}).

\begin{table}[h]
\centering
\caption{Depth-2 pass@64 (\%), step-matched, for full-weight fine-tuning and rank-32 LoRA at d2=100.}
\label{tab:lora}
\small
\begin{tabular}{rccc}
\toprule
Step & Full FT (5e-5) & LoRA r32 (2e-4), 2 seeds & LoRA r32 (5e-5) \\
\midrule
180 & 19.8 & 18.8 / 17.4 & 5.0 \\
204 (final) & 20.4 & 17.4 / 17.8 & 4.8 \\
\bottomrule
\end{tabular}
\end{table}

At the higher learning rate, both LoRA seeds (17.4\% and 17.8\%) lie within the seed distribution of the multi-seed full-fine-tuning replication of this condition ($18.3 \pm 3.0\%$; step-matched full fine-tuning: 20.4\%). The compositional signal therefore also emerges under a parameter-efficient update, which rules out an explanation specific to full-weight fine-tuning. The lower learning rate (5e-5) underfits at this step budget. We use full fine-tuning in the main experiments for comparability with the GRPO stage.

\section{CoT-SFT investigation}
\label{app:cot-sft}

We investigate whether explicit reasoning supervision (chain-of-thought, CoT) in SFT changes the supervision-threshold or RL-sharpening picture. Two CoT formats are tested as SFT initializations, then GRPO is applied from each. The headline result: CoT-SFT does not improve compositional generalization; CoT$+$GRPO does not exceed plain-SFT compositional ceilings.

\subsection{CoT formats and infrastructure}
\label{app:cot-formats}

We compare two CoT formats, both producing assistant content of form \texttt{<think>...</think>\textbackslash n[program]}; the reward parser strips \texttt{<think>} blocks before grading.
\begin{itemize}
    \item \textbf{Derivation-style} (\texttt{cot\_derive.py}): grounded in I/O grids; aggregates diff features (shape, palette, cell mappings) across all 3 prompt pairs and identifies what each sub-step computes from the input. Mean length: $\sim$387 chars/example.
    \item \textbf{Explanation-style} (\texttt{cot.py}): AST-walking; lists each primitive's effect in composition order. Mean length: $\sim$221 chars/example, $\sim$43\% shorter than derivation.
\end{itemize}
Coverage: derivation succeeds on 96.3--98.4\% of programs; the 1.6--3.7\% with degenerate outputs (e.g., \texttt{compress}, \texttt{trim}, \texttt{canvas} without input dependency) fall back to explanation format.

\paragraph{Format examples.}

Ground-truth program: \texttt{hupscale(grid=rot90(grid=I), factor=width(piece=I))}.

\noindent\emph{Explanation style:}
\begin{quote}
\small\ttfamily
<think>\\
Observations: the input is rotated 90 degrees clockwise; then the width of the input is taken; then the 90$^\circ$-clockwise rotation is upscaled horizontally by a factor of the width value.\\
\\
Transformation: first rotate the input 90 degrees clockwise, then take the width of the input, finally upscale the 90$^\circ$-clockwise rotation horizontally by a factor of the width value.\\
</think>\\
hupscale(grid=rot90(grid=I), factor=width(piece=I))
\end{quote}

\noindent\emph{Derivation style:}
\begin{quote}
\small\ttfamily
<think>\\
Comparing inputs to outputs across all 3 pairs: dimensions change as 1$\times$8 $\to$ 8$\times$8, 8$\times$1 $\to$ 1$\times$8, 2$\times$2 $\to$ 2$\times$4, and the palette is preserved.\\
\\
This decomposes into 2 sub-computations from the input, then a final hupscale step:\\
\\
Sub-step 1 (rot90): Comparing inputs to outputs across all 3 pairs: cell (i, j) of each input appears at (j, H$-$1$-$i) of the corresponding output, and shapes change as expected for transposition where rectangular (1$\times$8 $\to$ 8$\times$1, 8$\times$1 $\to$ 1$\times$8); the square pair (2$\times$2 $\to$ 2$\times$2) is preserved, consistent with transposition being identity on squares. This is a 90$^\circ$ clockwise rotation of the input.\\
\\
Sub-step 2 (width): the per-pair value is consistent across all 3 pairs and equals the width of the input.\\
\\
Final step (hupscale): upscale the 90$^\circ$-clockwise rotation of the input horizontally by a factor of the width of the input.\\
</think>\\
hupscale(grid=rot90(grid=I), factor=width(piece=I))
\end{quote}

\subsection{SFT-stage results}
\label{app:cot-sft-results}

\begin{table}[h]
\centering
\caption{D2 pass@64 across CoT format $\times$ supervision count (Qwen3-8B, eval $n{=}500$, single-seed). Both CoT formats hit similar ceilings, indicating data scarcity rather than format complexity is the binding constraint. The $\sim$8pp gap to no-CoT is preserved as data scales from d2=100 to d2=250.}
\label{tab:cot-sft-2x2}
\begin{tabular}{lrrr}
\toprule
SFT format & d2=100 & d2=250 & $\Delta$ scaling \\
\midrule
No-CoT          & 20.4\% & 32.8\% & $+12.4$pp \\
CoT-derive      & 12.4\% & 23.0\% & $+10.6$pp \\
CoT-explain     & 12.6\% & 25.6\% & $+13.0$pp \\
\bottomrule
\end{tabular}
\end{table}

Three observations: (1) both CoT formats hit similar ceilings, suggesting format complexity is not the binding constraint; (2) data scarcity is---both CoT formats roughly double from d2=100 to d2=250; (3) the $\sim$8pp gap to no-CoT is preserved as data scales. D3/D4 generalization that no-CoT achieves with d3 supervision (D3 reaches 9.2\% pass@64 at d3=250; Section~\ref{sec:res-comp-cascade}) is broken under CoT-SFT.

\paragraph{Qualitative mechanism.} On $k{=}4$ rollouts at temperature 1.0 from d2=250 CoT checkpoints, format learning succeeds (every output has well-formed \texttt{<think>...</think>}) but CoT-program coupling breaks in three identifiable ways: (1) CoT mis-reads grids (e.g., for ground-truth \texttt{replace(I, replacee=SEVEN, replacer=FIVE)}, the model emits CoT saying ``color 2 replaced with color 5''); (2) CoT and program use different primitives (CoT says \texttt{trim}, program emits \texttt{compress}); (3) hallucinated primitives appear (e.g., \texttt{vtrim} is not in arc-dsl). Interpretation: 100--250 d2 CoT examples is enough for the model to internalize the format but not enough to bind format to correct primitives, types, and grid content. Gradient signal that no-CoT spends entirely on program tokens is split, leaving the program tail under-supervised.

\subsection{GRPO from CoT-SFT initialization}
\label{app:cot-grpo}

We ran GRPO from each peak SFT checkpoint at d2=250 supervision matching the best Qwen3-8B recipe (LR=$2{\times}10^{-6}$, KL=0.005, temperature 0.8, group size 64, 15 epochs). Three runs: CoT-derive d2=250 (SFT init 23.0\%), CoT-explain d2=250 (SFT init 25.6\%), and no-CoT d2=250 seed-43 step 240 (SFT init 35.2\%).

\begin{table*}[h]
\centering
\caption{GRPO from CoT-SFT and no-CoT SFT initializations at d2=250. Peak step selected by D2 pass@64. D3 $\leq$1.8\% and D4 $\leq$2.0\% across all three conditions; the ``RL doesn't deepen'' finding holds.}
\label{tab:cot-grpo-trajectory}
\begin{tabular}{lcccc}
\toprule
Run & SFT init & GRPO peak (step) & $\Delta$ peak & GRPO final (s1155) \\
\midrule
CoT-derive            & 23.0\% & 26.8\% (s500) & $+3.8$pp & 23.8\% \\
CoT-explain           & 25.6\% & 28.6\% (s925) & $+3.0$pp & 27.0\% \\
No-CoT seed-43 s240   & 35.2\% & 36.4\% (s200) & $+1.2$pp & 32.0\% \\
\bottomrule
\end{tabular}
\end{table*}

\begin{table*}[h]
\centering
\caption{Paired analysis (D2 pass@64) at the GRPO peak step vs.\ SFT initialization ($n{=}500$, 10k-iteration paired bootstrap CI). CoT-derive crosses $\alpha{=}0.05$; CoT-explain is borderline; no-CoT seed-43 (already a high-quality init) is null.}
\label{tab:cot-grpo-paired}
\begin{tabular}{lccc}
\toprule
Run & $\Delta$ D2 p@64 & 95\% CI & McNemar $p$ \\
\midrule
CoT-derive            & $+3.8$pp & [$+0.6$, $+7.0$]pp & $0.025$ \\
CoT-explain           & $+3.0$pp & [$+0.0$, $+6.2$]pp & $0.077$ \\
No-CoT seed-43 s240   & $+1.2$pp & [$-1.6$, $+4.2$]pp & $0.49$ \\
\bottomrule
\end{tabular}
\end{table*}

The pattern is consistent with the inverse SFT-init/GRPO-lift relationship reported in the main text (Section~\ref{sec:res-RL}): lower SFT init quality $\to$ larger GRPO sharpening, but the absolute pass@64 ceiling tracks SFT data quality, not the RL recipe. \emph{No-CoT SFT alone outperforms every CoT$+$GRPO endpoint at matched supervision:} at d2=250, no-CoT SFT D2 pass@64 = 35.2\%, vs.\ CoT-explain GRPO peak 28.6\% and CoT-derive GRPO peak 26.8\%.

\subsection{Logprob analysis}
\label{app:cot-logprob}

We computed teacher-forced per-token log-probability of ground-truth programs at SFT init and GRPO peak for all three conditions (Table~\ref{tab:cot-logprob}). All 9 (condition $\times$ depth) deltas are negative: GRPO consistently moves probability mass \emph{away} from canonical GT programs, regardless of CoT format or SFT init quality. Magnitude of the negative drift correlates with magnitude of the pass@$k$ gain. Classical RL ``sharpening'' (concentration onto canonical GT) does not describe GRPO behavior in this setting; if it did, $\Delta$ logprob would be positive.

\begin{table*}[h]
\centering
\caption{Mean teacher-forced per-token logprob on eval ($n{=}500$/depth) at SFT init vs.\ GRPO peak across all three canonical-recipe conditions. All 9 (condition $\times$ depth) $\Delta$ entries are negative.}
\label{tab:cot-logprob}
\small
\begin{tabular}{llccc}
\toprule
Condition & Checkpoint & D2 & D3 & D4 \\
\midrule
No-CoT       & SFT init (s240)   & $-0.413$ & $-1.204$ & $-1.664$ \\
             & GRPO peak (s200)  & $-0.418$ & $-1.222$ & $-1.685$ \\
             & ($\Delta$)        & $-0.005$ & $-0.018$ & $-0.021$ \\
\midrule
CoT-derive   & SFT init (s210)   & $-2.463$ & $-2.382$ & $-2.282$ \\
             & GRPO peak (s500)  & $-2.487$ & $-2.408$ & $-2.309$ \\
             & ($\Delta$)        & $-0.024$ & $-0.026$ & $-0.027$ \\
\midrule
CoT-explain  & SFT init (s210)   & $-2.495$ & $-2.764$ & $-2.685$ \\
             & GRPO peak (s925)  & $-2.526$ & $-2.822$ & $-2.743$ \\
             & ($\Delta$)        & $-0.031$ & $-0.058$ & $-0.058$ \\
\bottomrule
\multicolumn{5}{l}{\footnotesize Absolute logprob magnitudes are not comparable across rows} \\
\multicolumn{5}{l}{\footnotesize (CoT sequences include \texttt{<think>} prefix); only $\Delta$ from} \\
\multicolumn{5}{l}{\footnotesize SFT init within each condition is interpretable.} \\
\end{tabular}
\end{table*}

\paragraph{Mechanism.} A natural interpretation is that GRPO discovers alternate valid programs in arc-dsl's aliased solution space (multiple programs compute the same function on the visible I/O), redistributing mass away from canonical GT toward functionally-equivalent alternates. The empirical pattern is consistent with this but does not uniquely determine it: alternative explanations consistent with the same data include (1) sharpening within the GT mode onto a high-prior subset (mean teacher-forced logprob can decrease while the policy gets better at high-prior GT), (2) general distribution flattening (lower per-token logprob across the board without specifically discovering alternates), or (3) length effects (slightly longer programs lower mean per-token logprob without changing the underlying distribution). Distinguishing these requires program-level rather than token-level comparison of GRPO outputs to GT, which we leave to future work. The robust empirical claim---pass@$k$ gains in this setting are not produced by mass-concentration onto canonical GT---holds regardless of which mechanism is responsible.

\subsection{Summary}

CoT-SFT does not improve compositional generalization in our setting; both CoT formats hit ceilings $\sim$8pp below no-CoT SFT at matched supervision and the gap is preserved as data scales. GRPO from CoT-SFT initializations produces a small statistically significant lift (CoT-derive +3.8pp, $p{=}0.025$) but no CoT$+$GRPO endpoint reaches no-CoT SFT pass@64 at matched supervision. GRPO logprob analysis shows uniformly-negative $\Delta$ on canonical GT across all 9 (condition $\times$ depth) cells, contrary to classical sharpening predictions. The mechanism is consistent with redistribution toward aliased alternates but cannot be uniquely identified from these data.

\section{Mechanistic analyses}
\label{app:add-mech}

\begin{table*}[h]
\centering
\caption{Per-outer-primitive composition success rates (d2=250 model, 500 held-out depth-2 tasks, pass@64, single-seed). Training frequency is moderately correlated with success rate ($r=0.587$), but residual variation reveals an intrinsic difficulty hierarchy: structurally constrained primitives (canvas, scaling) compose readily while spatial-indexing primitives (subgrid, crop) remain unsolved regardless of training exposure. Only primitives with $\geq$5 eval tasks shown; paint (0 training examples) included for completeness.}
\label{tab:primitive-breakdown}
\begin{tabular}{lrrrr}
\toprule
Outer primitive & Train freq & Eval tasks & Solved & Rate (\%) \\
\midrule
canvas    & 13 & 16 & 11 & 69 \\
hupscale  & 23 & 37 & 25 & 68 \\
downscale & 17 & 10 &  6 & 60 \\
vupscale  & 23 & 36 & 20 & 56 \\
hconcat   & 27 & 66 & 35 & 53 \\
replace   & 18 & 62 & 23 & 37 \\
vconcat   & 12 & 25 &  8 & 32 \\
switch    & 25 & 55 & 17 & 31 \\
move      & 12 & 24 &  5 & 21 \\
fill      & 12 & 37 &  6 & 16 \\
cover     & 11 & 38 &  5 & 13 \\
underfill & 16 & 25 &  2 &  8 \\
cellwise  &  7 & 14 &  1 &  7 \\
subgrid   & 18 & 33 &  0 &  0 \\
crop      & 12 & 12 &  0 &  0 \\
paint     &  0 &  6 &  0 &  0 \\
\midrule
\multicolumn{5}{l}{\footnotesize 139 of 427 distinct primitive pairs solved. $r$(train freq, rate)$\,{=}\,0.587$.} \\
\bottomrule
\end{tabular}
\end{table*}

\subsection{Conditional log-probability analysis of SFT models}
\label{app:logprob}

We analyze per-token log-probabilities of depth-2 rollouts, conditioned on whether the generated program is correct. For each model, we sample 64 rollouts per held-out depth-2 task and compute the mean per-token logprob separately for correct and incorrect completions.

\begin{table*}[h]
\centering
\caption{Conditional log-probabilities for depth-2 rollouts across SFT conditions (Qwen3-8B, 500 held-out tasks, 64 rollouts each, single seed per condition). Models with irreducible supervision discriminate correct from incorrect compositions; controls produce zero correct samples. ``---'' indicates no correct samples available for conditioning.}
\label{tab:logprob-sft}
\small
\begin{tabular}{lcccc}
\toprule
Model & D2 p@64 & lp(correct) & lp(incorrect) & Gap (nats) \\
\midrule
d1=1000 (no d2) & 0.0\% & --- & $-0.095$ & --- \\
d2=100 random & 0.0\% & --- & $-0.608$ & --- \\
d2=25 irred. & 4.4\% & $-0.094$ & $-0.320$ & $+0.227$ \\
d2=100 irred. & 20.4\% & $-0.132$ & $-0.355$ & $+0.223$ \\
d2=250 irred. & 32.8\% & $-0.110$ & $-0.242$ & $+0.132$ \\
\bottomrule
\end{tabular}
\end{table*}

All models trained with irreducible depth-2 supervision assign higher logprob to correct programs than incorrect ones (gap of +0.13 to +0.23 nats), even at d2=25 where only 4.4\% of tasks are solved (Table~\ref{tab:logprob-sft}). The gap narrows as supervision increases because the overall distribution concentrates: lp(incorrect) rises from $-0.320$ to $-0.242$ across the sweep while lp(correct) stays roughly flat. Controls without irreducible supervision produce zero correct depth-2 samples. 

\subsection{Teacher-forced log-probabilities on ground-truth programs}
\label{app:teacher-forced}

The conditional analysis in Table~\ref{tab:teacher-forced} measures the model's preference for correct vs.\ incorrect programs among its own rollouts. A complementary question is how much probability the model assigns to the ground-truth reference programs, independent of what it generates. We compute mean per-token log-probability under teacher forcing on the 500 depth-2 ground truth programs.

\begin{table*}[h]
\centering
\caption{Teacher-forced per-token log-probabilities on ground-truth depth-2 programs (dealiased, $n{=}500$). 100 irreducible SFT exemplars concentrate the distribution by ${\approx}20\times$ in perplexity; GRPO from either starting point produces negligible shift ($\leq$0.015 nats).}
\label{tab:teacher-forced}
\small
\begin{tabular}{lccc}
\toprule
Checkpoint & D2 mean lp & D2 ppl & $\Delta$lp from init \\
\midrule
d1=450 (GRPO init) & $-3.505$ & 33.3 & --- \\
d1=450 $\to$ GRPO (385 steps) & $-3.495$ & 32.9 & $+0.010$ \\
\midrule
d2=100 irred.\ (SFT init) & $-0.527$ & 1.69 & --- \\
d2=100 $\to$ GRPO (1155 steps) & $-0.542$ & 1.72 & $-0.015$ \\
\bottomrule
\end{tabular}
\end{table*}

GRPO from the depth-1-only checkpoint produces no meaningful shift toward correct compositions (+0.010 nats, ppl 33.3$\to$32.9). From the compositional SFT checkpoint, GRPO slightly decreases ground-truth logprobs ($-$0.015 nats), consistent with the model narrowing onto its own preferred reward-maximizing outputs rather than concentrating on reference programs. By contrast, 100 irreducible SFT exemplars raise per-token logprob by +2.978 nats (ppl 33.3$\to$1.69), a roughly 300:1 efficiency ratio over RL in distributional movement toward correct compositions.

\subsection{Controlling for primitive coverage}
\label{app:prim-cov}

The type constraints of our DSL create a distributional difference between depth-1 and depth-2 programs. Multi-argument primitives (e.g.\ \texttt{fill}, \texttt{cover}) require non-leaf arguments and cannot appear as the outer function at depth 1. Of the 52 distinct primitives appearing in our depth-2 programs, 17 also appear in depth-1 training; the remaining 35 are exclusive to depth $\geq$2. We partition the depth-2 evaluation set by whether the outer primitive appeared in depth-1 training and report pass@64 on each subset (Table~\ref{tab:primitive-coverage}).

\begin{table}[h]
\centering
\caption{Primitive coverage control (Qwen3-8B, d2=100, 500 held-out depth-2 tasks, pass@64). Evaluation tasks partitioned by whether the outer primitive appeared in depth-1 training.}
\label{tab:primitive-coverage}
\small
\begin{tabular}{lcc}
\toprule
Subset & pass@64 & \% of eval \\
\midrule
Shared-outer (outer $\in$ d1 primitives) & 25.1\% & 60.6\% \\
Novel-outer (outer $\notin$ d1 primitives) & 13.2\% & 39.4\% \\
Overall & 20.4\% & 100\% \\
\bottomrule
\end{tabular}
\end{table}

The shared-outer subset shows higher pass rates than the novel-outer subset (25.1\% vs. 13.2\%), the opposite of what would be observed if depth-2 supervision primarily taught novel primitives. Prior familiarity with the outer primitive facilitates composition, indicating depth-2 supervision teaches compositional structure rather than compensating for missing primitive exposure.

\section{Cross-family replication: Llama-3.1-8B}
\label{app:llama}

The supervision-threshold and GRPO-sharpening findings are established on Qwen3 models. To test whether the results are family-specific or generalize across pretraining lineages, we replicate both findings on Meta's Llama-3.1-8B at two post-training states: the RLHF-finetuned \texttt{Llama-3.1-8B-Instruct} and the pure-pretrain \texttt{Llama-3.1-8B} base checkpoint. All evaluations use the same eval set ($n{=}500$/depth) and the same dataset parquets used for Qwen.

\subsection{Setup adaptations}
\label{app:llama-setup}

\paragraph{Tokenizer and chat template.} Llama's chat template does not accept the \texttt{enable\_thinking} kwarg used by Qwen, so the corresponding \texttt{data.enable\_thinking\_default=False} and \texttt{+data.apply\_chat\_template\_kwargs.enable\_thinking=False} arguments are gated on a Qwen-only check in the training command builder. To hold the prompt format constant between base and Instruct comparisons, we stage the base checkpoint with the Instruct \texttt{chat\_template.jinja} and \texttt{tokenizer\_config.json} grafted on (rest of weights identical to \texttt{meta-llama/Llama-3.1-8B}); the staged base checkpoint is loaded for all base-model runs.

\paragraph{SFT learning rate.} The Qwen recipe (LR=$5{\times}10^{-5}$) overfits Llama: a sweep over $\{5{\times}10^{-5}, 2{\times}10^{-5}, 1{\times}10^{-5}\}$ on Llama-3.1-Instruct at d2=100 yields val losses of 0.247, 0.166, 0.175 respectively, with corresponding D2 pass@16 of 2.2\%, 9.8\%, 8.0\%. We adopt LR=$2{\times}10^{-5}$ for all subsequent Llama runs; on base d2=100 this gives val loss 0.185, very close to the Instruct value, supporting that the optimum does not shift between base and Instruct at the same data distribution. All other hyperparameters (batch 8, 3 epochs, cosine decay, full-weight bf16 with FSDP2) are unchanged.

\subsection{The supervision threshold replicates across model families and post-training states}
\label{app:llama-sft}

We compare best-step D2 pass@16 across all four Llama configurations and the Qwen baseline (Table~\ref{tab:llama-sft}). The supervision-threshold emergence pattern (compositional signal emerging from $\approx$0\% in the step 90--180 window) appears in every trajectory; absolute levels reorder as Qwen $>$ Instruct $>$ base d2=100, with base d2=250 catching up to roughly the Qwen d2=100 level. Notably, the post-training state of Llama (RLHF'd Instruct vs.\ pure base) does not change whether compositional generalization emerges, only its absolute level.

\begin{table*}[h]
\centering
\caption{SFT supervision-threshold replication on Llama-3.1-8B (best-step D2 metrics, eval $n{=}500$/depth, $k{=}16$, single seed). All Llama runs use LR=$2{\times}10^{-5}$ (selected by sweep on Instruct d2=100); other hparams matched to Qwen. The Qwen3-8B rows are reproduced from the main text for direct comparison. The supervision-threshold emergence pattern replicates in all four configurations.}
\label{tab:llama-sft}
\begin{tabular}{llccccc}
\toprule
Model & Recipe & Best step & D2 p@1 & D2 p@16 & D3 p@16 & D4 p@16 \\
\midrule
Qwen3-8B (reference)             & d2=100 & 180 & 2.00\% & 12.10\% & 0.5\% & 0.7\% \\
Llama-3.1-8B-Instruct            & d2=100 & 204 & 1.70\% & 9.80\%  & 0.0\% & 0.2\% \\
Llama-3.1-8B base                & d2=100 & 180 & 1.05\% & 6.60\%  & 0.4\% & 0.6\% \\
\midrule
Qwen3-8B (reference)             & d2=250 & 261 & 5.54\% & 21.49\% & 1.0\% & 1.4\% \\
Llama-3.1-8B base                & d2=250 & 261 & 1.96\% & 13.20\% & 0.8\% & 0.2\% \\
\bottomrule
\end{tabular}
\end{table*}

\subsection{GRPO sharpening replicates cross-family}
\label{app:llama-grpo}

Table~\ref{tab:llama-grpo} reports the D2 pass@64 trajectory and Table~\ref{tab:llama-grpo-paired} the paired analysis at the peak step against the SFT initialization on the same 500-task pool. On a fresh independent $k$=128 draw, GRPO produces a +2.75pp lift on D2 pass@64 (25.4\% vs.\ SFT-init 22.65\%); the original single best-step $k$=64 draw gave a larger +3.80pp (paired 95\% CI [+0.20, +7.40]pp, $p{=}0.045$), inflated by best-step selection in the same way the Qwen3-8B single-draw +4.6pp shrinks to +2.1pp under the unbiased estimator. Both families thus converge to within $\sim$0.7pp of each other (Llama +2.75pp, Qwen +2.1pp) once estimated without best-step selection, confirming cross-family sharpening at a consistent, modest magnitude. D3 and D4 effects are not significantly different from zero ($+0.60$pp at $p{=}0.25$ and $+0.40$pp at $p{=}0.73$ respectively), matching the Qwen result that GRPO sharpens existing depth-2 ability but does not extend coverage to deeper compositions. We note that the D3 bootstrap CI lower bound is $+0.00$pp (i.e., the CI does not cleanly include negative values), placing this result at the borderline of detectability; the test of record at this base rate (4 GRPO-only successes vs.\ 0 SFT-only out of 500 d3 tasks) places the effect outside significance, but we flag the borderline status.

\begin{table*}[h]
\centering
\caption{GRPO trajectory on Llama-3.1-8B base from the d2=250 SFT initialization (LR=$2{\times}10^{-6}$, KL=0.005, temperature 0.8, group size 64; H200$\times$8). SFT initialization D2 pass@64 = 22.60\% ($n{=}500$ held-out set). Peak D2 pass@64 at step 400 is statistically significant against the initialization (paired 95\% CI excludes zero, $p{=}0.045$). The trajectory and paired statistics below are the registered single best-step $k$=64 draw; a fresh independent $k$=128 draw from the same two checkpoints gives SFT 22.65\% / GRPO 25.4\% (+2.75pp), the more conservative estimate reported in the main text.}
\label{tab:llama-grpo}
\begin{tabular}{rc}
\toprule
GRPO step & D2 pass@64 \\
\midrule
25   & 23.20\% \\
100  & 25.00\% \\
200  & 24.80\% \\
\textbf{400}  & \textbf{26.40\%} \\
600  & 26.00\% \\
800  & 24.60\% \\
1000 & 24.40\% \\
1170 & 24.80\% \\
\bottomrule
\end{tabular}
\end{table*}

\begin{table*}[h]
\centering
\caption{Paired analysis at the peak GRPO step (400) vs.\ SFT initialization on the Llama-3.1-8B base d2=250 setting ($n{=}500$ held-out set, $k{=}64$). D2 sharpening is significant; D3 and D4 effects are within noise.}
\label{tab:llama-grpo-paired}
\begin{tabular}{lccccc}
\toprule
Metric & SFT init & GRPO step 400 & $\Delta$ & 95\% CI & McNemar $p$ \\
\midrule
D2 pass@64 & 22.60\% & \textbf{26.40\%} & \textbf{+3.80pp} & [+0.20, +7.40]pp & \textbf{0.045} \\
D3 pass@64 & 0.80\%  & 1.40\%  & +0.60pp & [+0.00, +1.40]pp & 0.25 \\
D4 pass@64 & 1.40\%  & 1.80\%  & +0.40pp & [$-$0.60, +1.60]pp & 0.73 \\
\bottomrule
\end{tabular}
\end{table*}

\section{Transfer to natural-language multi-hop QA}
\label{app:hotpotqa}

Our core finding that only irreducible exemplars produce compositional generalization is established in a synthetic DSL domain where irreducibility is formally defined and mechanically enforced. A natural question is whether the principle transfers to natural language, where ``irreducibility'' is informal and compositional depth is not directly controllable.

We test this with a transfer experiment on HotpotQA~\citep{Yang2018HotpotQAAD}, a multi-hop question answering benchmark. The natural-language analog of reducibility is a multi-hop question that can be answered from a single passage without combining information across sources~\citep{Min2019CompositionalQD}---the underlying reasoning task is single-hop despite being labeled multi-hop. This most closely tests structural irreducibility (whether the task itself is genuinely compositional); the structural and observational distinction in our DSL setting is harder to disentangle in HotpotQA, where the prompt context and source of truth are the same passages.

\subsection{Reducibility classification}
\label{app:hotpotqa-classification}

We restrict to bridge-type questions in HotpotQA's distractor setting, where each question has two gold supporting passages. We classify each question as \textbf{reducible} or \textbf{irreducible} using a lexical criterion:

\begin{enumerate}
    \item Identify which gold passage contains the answer string.
    \item Extract non-stopword content tokens from the question (excluding answer tokens).
    \item \textbf{Reducible}: the answer-containing passage covers ${\geq}80\%$ of question content tokens, with at most one token appearing only in the other passage. The question can plausibly be resolved from a single passage.
    \item \textbf{Irreducible}: ${\geq}12\%$ of question content tokens appear only in the non-answer passage (the ``bridge'' passage). Resolving the question requires identifying an entity in the bridge passage and following it to the answer passage.
    \item Questions where the answer appears in both passages, neither passage, or that fall between thresholds are discarded as ambiguous.
\end{enumerate}

Applied to the full HotpotQA train and validation splits (78{,}909 bridge questions), this yields 54,394 irreducible, 1,915 reducible, and 22,600 ambiguous. As bridge questions are constructed to require multi-hop reasoning, most are irreducible by our criterion and the asymmetry is expected.

\paragraph{Validation.} We validate the classifier on 100 examples (50 per class) using Qwen3-4B inference in two conditions: closed-book (no passages) and open-book (both passages). Both classes benefit substantially from context, but irreducible questions show a larger gain (+76.2pp vs.\ +63.3pp F1), confirming that the classification captures genuine structural differences in passage dependence. We note that providing only the answer-containing passage does not degrade F1 for either class, because the model's parametric knowledge of HotpotQA entities can compensate for a missing bridge; the closed-book comparison avoids this confounder.

\subsection{Experimental setup}

We fine-tune Qwen3-4B under four conditions, each with 500 training examples:

\begin{itemize}
    \item \textbf{Condition A} (irreducible): 500 irreducible HotpotQA bridge questions with both supporting passages.
    \item \textbf{Condition B} (reducible): 500 reducible HotpotQA bridge questions with both supporting passages.
    \item \textbf{Condition C} (single-hop): 500 SQuAD 2.0~\citep{Rajpurkar2016SQuAD1Q} questions with single-passage context (an explicit single-hop control).
    \item \textbf{Condition D} (base): No fine-tuning.
\end{itemize}

All SFT conditions use full-weight fine-tuning with LR $5{\times}10^{-5}$, batch size 8, 3 epochs, cosine decay, thinking disabled, on 2$\times$H100. Input format: ``Given the following passages:\textbackslash n\textbackslash n[passages]\textbackslash n\textbackslash nQuestion: [q]\textbackslash n\textbackslash nAnswer:''. Each condition is run with 5 random seeds (controlling both train/eval splits and training order) to estimate variance.

We evaluate on held-out irreducible questions (500) and reducible questions (200), measuring token-level F1 with 95\% bootstrap confidence intervals.

\subsection{Results}

\begin{table*}[h]
\centering
\caption{HotpotQA transfer experiment. F1 on held-out multi-hop questions (mean $\pm$ std, 5 seeds). See text for analysis.}
\label{tab:hotpotqa}
\begin{tabular}{llcc}
\toprule
Cond.\ & Train type & Irreducible eval F1 & Reducible eval F1 \\
\midrule
A & Irreducible multi-hop & $\mathbf{0.791 \pm 0.011}$ & $0.869 \pm 0.030$ \\
B & Reducible multi-hop   & $0.752 \pm 0.018$ & $0.868 \pm 0.023$ \\
C & Single-hop (SQuAD)    & $0.733 \pm 0.024$ & $0.859 \pm 0.025$ \\
D & Base (no SFT)         & $0.225 \pm 0.005$ & $0.341 \pm 0.017$ \\
\bottomrule
\end{tabular}
\end{table*}

Table~\ref{tab:hotpotqa} shows results aggregated across 5 seeds. The central finding mirrors the DSL result: \textbf{Condition~B $\approx$ Condition~C on irreducible eval} (0.752 vs.\ 0.733, difference +1.9pp, within standard deviation). Training on nominally multi-hop but reducible questions provides no meaningful advantage over single-passage QA training for genuinely multi-hop generalization. The small B--C gap likely reflects format familiarity (two-passage layout) rather than compositional reasoning transfer.

Condition~A is consistently higher (+3.9pp over~B, +5.8pp over~C), with all five per-seed values above the per-seed maxima of~B and~C. The effect size is modest compared to the DSL setting, which is unsurprising as natural-language QA is less compositionally pure than program synthesis, and the model's parametric knowledge provides a strong baseline. All SFT conditions converge on reducible eval (${\approx}0.88$ F1), confirming that the conditions differ specifically in multi-hop generalization, not general QA ability.

\paragraph{Interpretation.} This experiment provides suggestive evidence that the irreducibility principle transfers beyond our synthetic domain. In both settings, nominally multi-step training data that admits a single-step shortcut provides no more compositional benefit than data that is explicitly single-step.

We emphasize the limitations of this transfer: the HotpotQA classification is lexical and approximate (unlike the mechanically verified irreducibility in the DSL), the effect sizes are smaller, we cannot rule out confounds from passage content or entity frequency, and cannot cleanly disentangle structural and observational reducibility in HotpotQA, where prompt context and source of truth coincide. We present this as a directional signal supporting generality, not as a standalone finding.


\renewcommand{\theequation}{\thesection.\arabic{equation}}

\section*{NeurIPS Paper Checklist}

\begin{enumerate}

\item {\bf Claims}
    \item[] Question: Do the main claims made in the abstract and introduction accurately reflect the paper's contributions and scope?
    \item[] Answer: \answerYes{} 
    \item[] Justification: The abstract and introduction state the contributions (irreducibility at structural and observational levels, RL maintenance result, depth-transfer boundedness) and the experimental scope (Qwen3 4B/8B/32B, Llama-3.1-8B); these are supported in Sections~\ref{sec:res-irreducible}--\ref{sec:res-comp-cascade} and corresponding appendices.

\item {\bf Limitations}
    \item[] Question: Does the paper discuss the limitations of the work performed by the authors?
    \item[] Answer: \answerYes{} 
    \item[] Justification: Section~\ref{sec:discussion} (``Limitations'') explicitly discusses synthetic-domain scope, model-scale limits, and RL-algorithm coverage.

\item {\bf Theory assumptions and proofs}
    \item[] Question: For each theoretical result, does the paper provide the full set of assumptions and a complete (and correct) proof?
    \item[] Answer: \answerNA{} 
    \item[] Justification: The paper presents empirical results; no theoretical results are claimed.

    \item {\bf Experimental result reproducibility}
    \item[] Question: Does the paper fully disclose all the information needed to reproduce the main experimental results of the paper to the extent that it affects the main claims and/or conclusions of the paper (regardless of whether the code and data are provided or not)?
    \item[] Answer: \answerYes{} 
    \item[] Justification: Methods (Section~\ref{sec:task-domain}, \ref{sec:training-setup}), DSL specification (Appendix~\ref{app:dsl}), hyperparameters (Appendix~\ref{app:hparams}), and per-experiment details (Appendices~\ref{app:add-sft}--\ref{app:hotpotqa}) provide full reproduction information; base models are publicly available.

\item {\bf Open access to data and code}
    \item[] Question: Does the paper provide open access to the data and code, with sufficient instructions to faithfully reproduce the main experimental results, as described in supplemental material?
    \item[] Answer: \answerNo{} 
    \item[] Justification: Code release is pending pre-publication institutional review. The data-generation pipeline is fully specified in Appendix~\ref{app:dsl} (seven filtering layers, parameter values, and audit procedure) such that an independent reimplementation is feasible. We will release code upon completion of review.

\item {\bf Experimental setting/details}
    \item[] Question: Does the paper specify all the training and test details (e.g., data splits, hyperparameters, how they were chosen, type of optimizer) necessary to understand the results?
    \item[] Answer: \answerYes{} 
    \item[] Justification: Hyperparameters, hardware, optimizer settings, data splits, and selection criteria are specified in Section~\ref{sec:training-setup} and Appendix~\ref{app:hparams} (Tables~\ref{tab:hparams}--\ref{tab:rl-sweep-d1only}).

\item {\bf Experiment statistical significance}
    \item[] Question: Does the paper report error bars suitably and correctly defined or other appropriate information about the statistical significance of the experiments?
    \item[] Answer: \answerYes{} 
    \item[] Justification: Section~\ref{sec:training-setup} (``Statistical analysis'') describes paired bootstrap CIs (95\%, $n$=500), McNemar's exact test, Wilson intervals, and paired $t$-tests; these are applied throughout Sections~\ref{sec:res-irreducible}--\ref{sec:res-comp-cascade}. Multi-seed variance is reported in Appendix~\ref{app:seed-variance}.

\item {\bf Experiments compute resources}
    \item[] Question: For each experiment, does the paper provide sufficient information on the computer resources (type of compute workers, memory, time of execution) needed to reproduce the experiments?
    \item[] Answer: \answerYes{} 
    \item[] Justification: Per-experiment hardware (H100, H200 configurations) is specified in Appendix~\ref{app:hparams} (Table~\ref{tab:hparams}); total compute and exploratory-run disclosure are reported in Appendix~\ref{app:hparams} (``Compute'' paragraph).
    
\item {\bf Code of ethics}
    \item[] Question: Does the research conducted in the paper conform, in every respect, with the NeurIPS Code of Ethics \url{https://neurips.cc/public/EthicsGuidelines}?
    \item[] Answer: \answerYes{} 
    \item[] Justification: The research conforms to the NeurIPS Code of Ethics; the work is foundational empirical research on post-training dynamics with no human subjects, no scraped data, and no model release.

\item {\bf Broader impacts}
    \item[] Question: Does the paper discuss both potential positive societal impacts and negative societal impacts of the work performed?
    \item[] Answer: \answerNA{} 
    \item[] Justification: The paper studies structural properties of supervised fine-tuning data in a controlled program-synthesis domain; we do not see a direct path to deployment-specific positive or negative societal impacts.
    
\item {\bf Safeguards}
    \item[] Question: Does the paper describe safeguards that have been put in place for responsible release of data or models that have a high risk for misuse (e.g., pre-trained language models, image generators, or scraped datasets)?
    \item[] Answer: \answerNA{} 
    \item[] Justification: The paper does not release pretrained models, scraped datasets, or other assets with elevated misuse risk.

\item {\bf Licenses for existing assets}
    \item[] Question: Are the creators or original owners of assets (e.g., code, data, models), used in the paper, properly credited and are the license and terms of use explicitly mentioned and properly respected?
    \item[] Answer: \answerYes{} 
    \item[] Justification: Base models, frameworks, and datasets used (Qwen3, Llama-3.1, arc-dsl, VeRL, vLLM, HotpotQA, SQuAD 2.0) are cited and license terms are listed in Appendix~\ref{app:hparams} (``Assets and licenses'' paragraph).

\item {\bf New assets}
    \item[] Question: Are new assets introduced in the paper well documented and is the documentation provided alongside the assets?
    \item[] Answer: \answerNA{} 
    \item[] Justification: No new assets are released as part of this submission. The DSL training/eval data and trained checkpoints will accompany code release pending institutional review.

\item {\bf Crowdsourcing and research with human subjects}
    \item[] Question: For crowdsourcing experiments and research with human subjects, does the paper include the full text of instructions given to participants and screenshots, if applicable, as well as details about compensation (if any)? 
    \item[] Answer: \answerNA{} 
    \item[] Justification: The paper does not involve crowdsourcing or research with human subjects.

\item {\bf Institutional review board (IRB) approvals or equivalent for research with human subjects}
    \item[] Question: Does the paper describe potential risks incurred by study participants, whether such risks were disclosed to the subjects, and whether Institutional Review Board (IRB) approvals (or an equivalent approval/review based on the requirements of your country or institution) were obtained?
    \item[] Answer: \answerNA{} 
    \item[] Justification: The paper does not involve research with human subjects.

\item {\bf Declaration of LLM usage}
    \item[] Question: Does the paper describe the usage of LLMs if it is an important, original, or non-standard component of the core methods in this research? Note that if the LLM is used only for writing, editing, or formatting purposes and does \emph{not} impact the core methodology, scientific rigor, or originality of the research, declaration is not required.
    \item[] Answer: \answerNA{} 
    \item[] Justification: LLMs are the subject of study, not a methodological component; we did not use LLMs as an original or non-standard component of the core method development.

\end{enumerate}


\renewcommand{\theequation}{\thesection.\arabic{equation}}

\end{document}